\ifdefined\XeTeXversion\else
  \pdfoutput=1
\fi
\documentclass[10pt]{article}
\usepackage[accepted]{tmlr}

\usepackage{amsmath,amssymb}
\usepackage{graphicx}
\usepackage{booktabs}
\usepackage{float}
\usepackage[hidelinks]{hyperref}
\usepackage{url}
\usepackage{listings}
\usepackage{xcolor}
\usepackage{tikz}
\usetikzlibrary{arrows.meta,positioning,shapes.geometric,fit,backgrounds,calc}

\newcounter{algorithm}

\lstdefinestyle{python}{
  language=Python,
  basicstyle=\ttfamily\scriptsize,
  keywordstyle=\color{blue},
  commentstyle=\color{gray},
  stringstyle=\color{red!60!black},
  showstringspaces=false,
  breaklines=true,
  frame=single,
  numbers=left,
  numberstyle=\tiny\color{gray},
}

\title{When LLM Reward Design Fails:\\Diagnostic-Driven Refinement for Sparse Structured RL}

\author{
  \name Youting Wang \email ginkoin613@gmail.com \\
  \addr Independent Researcher
  \AND
  \name Yuan Tang \email naomitangyuan@gmail.com \\
  \addr Independent Researcher
  \AND
  \name Bowen Liu \email chris3494285@gmail.com \\
  \addr Independent Researcher
  \AND
  \name Xuan Liu \email daisyxliuapp@gmail.com \\
  \addr Independent Researcher
  \AND
  \name Dingyan Shang \email dingyanshang@gmail.com \\
  \addr Independent Researcher
}

\def\month{09}
\def\year{2026}
\def\openreview{\url{https://openreview.net/forum?id=5TmwBBn8Oh}}

\begin{document}

\maketitle

\begin{abstract}
For sparse, structured reinforcement-learning tasks with semantic reward-function interfaces, LLM-generated reward shaping is better framed as a debugging problem than as pure one-shot generation. We study PPO-trained agents, using MiniGrid as the core sparse-structured evaluation and MuJoCo reaching/locomotion as boundary stress tests. Our qualitative audit finds two recurring one-shot failure modes---reward flooding and semantic/API misunderstanding---plus a rarer, less reliably distinguished weak-shaping case. We propose \textbf{diagnostic-driven iterative refinement}, where training diagnostics and a failure-mode taxonomy guide targeted reward-function revision. On sparse structured tasks with diagnosable reward failures, refinement improves DoorKey-8$\times$8 from $2.3\%$ success without shaping to $97.6\%$, and KeyCorridor from $31.2\%$ one-shot to $86.7\%$ with high seed-to-seed variance. Controls indicate that these gains are not attributable to generic retrying or extra training alone: metrics-only re-prompting produces large drops, while a static-vocabulary control recovers much of the gap, showing that the taxonomy prompt itself is a major mechanism and that dynamic trigger labels provide only partially isolated incremental evidence. In a new prompt-matched comparison across Haiku~4.5, Opus~5, and DeepSeek V4 Pro, the refinement effect is task- and model-dependent and none of the six contrasts survives family-wise correction; a four-call, episode-budget-matched population control is also competitive. Component-removal stress tests and sensitivity analyses reveal calibration limits. Continuous-control results expose the boundary: success-based diagnostics can misfire in dense-reward locomotion, and return-trend feedback removes one false-positive mechanism without producing robust locomotion gains. The method's scope is deliberately bounded to sparse structured tasks with reliable structured interfaces under PPO training.
\end{abstract}

\section{Introduction}
\label{sec:intro}

Reinforcement learning in environments with sparse rewards remains a fundamental challenge. An agent exploring randomly in a multi-step task---such as finding a key, unlocking a door, and reaching a goal---may never encounter the terminal reward signal within feasible training budgets. Reward shaping~\citep{ng1999policy} addresses this by providing intermediate feedback, but designing effective shaping functions demands substantial domain expertise and iterative manual tuning.

Recent work has explored LLMs as reward function generators: Eureka performs evolutionary optimization over reward code with GPU-parallel simulation, while Text2Reward generates dense reward code from language and compact environment representations~\citep{ma2023eureka,xie2023text2reward}. These systems achieve impressive results, but they provide limited insight into \emph{why} LLM-generated rewards fail---treating reward generation primarily as a performance-driven search or refinement problem rather than analyzing the structure of failures.

We argue that LLM reward design for sparse, structured tasks is better understood as a \emph{debugging} problem than a generation problem. One-shot LLM generation fails in recurring, identifiable ways; lightweight diagnostics can sometimes detect and repair these failures in the studied sparse structured setting; and the approach breaks predictably when diagnostics are misaligned with the task's evaluation signal. This reframing leads to a practical method---diagnostic-driven iterative refinement---that uses 1--4 LLM calls per reward-design run, repeated across seeds for evaluation, and CPU training. We evaluate the end-to-end protocol alongside prompt-matched controls that separate static failure-aware guidance from corrective refinement.

Our \textbf{main contribution} is a diagnostic-driven framework for identifying and repairing systematic failure modes in LLM-generated reward shaping, supported by targeted experiments that test the framework's core assumptions.
We provide three \textbf{supporting findings}:

\begin{enumerate}
  \item \textbf{A failure taxonomy for structured reward interfaces}: prompt controls show that taxonomy-rich feedback is functionally important, but also that static taxonomy vocabulary explains a substantial part of the improvement, so automated trigger labels should not be read as the sole causal mechanism. The taxonomy itself is grounded in a qualitative audit where reward flooding and semantic/API misunderstanding are the two dominant failure patterns, while weak or negligible shaping appears as a rarer and less reliably distinguished case.

  \item \textbf{Variance-based evidence} for when refinement helps: LLM generation variance has the clearest dominant point estimate in representative sparse tasks (DoorKey-8$\times$8: LLM std $47.5\%$ vs.\ RL std $0.3\%$; crossed LLM share $96\%$ with a wide $[66,100]\%$ bootstrap interval), making diagnostic refinement most useful in this regime; residual and RL-training variance have larger point estimates in representative continuous-control decompositions, limiting the benefit of reward-function correction.

  \item \textbf{A domain-boundary stress test}: success-based diagnostics produce false positives on dense-reward locomotion, systematically degrading the feedback loop. We provide root-cause analysis and evaluate a return-trend diagnostic adaptation, showing both where the framework breaks and how one specific failure mechanism can be removed.
\end{enumerate}

The prompt-matched evaluation further tests whether refinement remains useful when one-shot generation receives the same API description and static failure-mode guidance. Across three backbones and two tasks, the effects are task- and model-dependent and none of the six contrasts survives family-wise correction; a matched four-call population control is also competitive.

The method is strongest for sparse, structured tasks with reliable semantic state fields and exposed subgoal structure, and all experiments use PPO-trained policies. Richer natural-language event descriptions are not uniformly beneficial: an interface ablation shows that removing \texttt{event\_text} is neutral on DoorKey-8$\times$8 and improves a KeyCorridor rerun. We explicitly characterize this scope: it is not yet a claim about RLHF, open-ended LLM agents, raw-pixel reward design, arbitrary robotics tasks, or algorithm-agnostic reward-function validity, and dense-reward settings require different diagnostics (Section~\ref{sec:limitations}).

\textbf{Roadmap and evidence status.}
Table~\ref{tab:main_results} provides the headline 10-seed MiniGrid and reaching results; Section~\ref{sec:diagnostics_validation} tests the diagnostic story through lower-seed stress tests and prompt controls; Sections~\ref{sec:backbone_matched}--\ref{sec:bestofn} report the matched-backbone, boundary, budget, variance, and selection analyses. The central positive claim is about sparse structured PPO tasks with diagnosable reward-function failures.

\section{Background}
\label{sec:background}

\subsection{Reward Shaping}

Potential-based reward shaping~\citep{ng1999policy} provides a principled framework for adding intermediate rewards without altering the optimal policy. In practice, most reward shaping uses heuristic bonuses tied to task-specific subgoals (e.g., +0.2 for picking up a key). While effective, these heuristics require domain knowledge and manual tuning.

\subsection{LLMs for Code Generation in RL}

Eureka~\citep{ma2023eureka} demonstrated that LLMs can generate reward functions for continuous-control tasks, using evolutionary search over hundreds of candidates evaluated in GPU-parallel simulation (IsaacGym). Text2Reward~\citep{xie2023text2reward} generates dense executable reward code from language task descriptions and compact environment representations. Our work takes a complementary, diagnostic-driven approach: rather than searching broadly, we analyze \emph{why} generated rewards fail and use targeted feedback to repair specific failure modes with minimal iteration. We evaluate primarily on discrete MiniGrid tasks and include continuous MuJoCo tasks as stress tests for the diagnostic assumptions.

\subsection{MiniGrid Environments}

We use the MiniGrid framework~\citep{minigrid}, which provides a suite of grid-world environments with partial observability. The agent receives a $7 \times 7 \times 3$ symbolic observation encoding object type, color, and state for each visible cell. Actions are discrete: turn left, turn right, move forward, pick up, drop, toggle, and done.

\subsection{MuJoCo Continuous Control}

To probe the boundary beyond discrete grid worlds, we also evaluate on MuJoCo~\citep{todorov2012mujoco} continuous-control tasks via Gymnasium~\citep{towers2023gymnasium}. For reaching tasks: \textbf{Reacher-v4} (2-DOF planar arm, 11-dim observation, 2-dim action) and \textbf{FetchReach-v4} (7-DOF robotic arm, 13-dim observation, 4-dim action)~\citep{plappert2018multigoal}. For the reaching diagnostics, we treat success as an end-effector distance below $0.05$.

We further evaluate on standard locomotion benchmarks: \textbf{HalfCheetah-v4} (2D runner, 17-dim observation, 6-dim action) and \textbf{Hopper-v4} (single-legged hopper, 11-dim observation, 3-dim action). Unlike reaching tasks with sparse binary success, these feature dense multi-component rewards (forward velocity $+$ alive bonus $-$ control cost), making them a boundary test for diagnostics designed around binary success signals.

\subsection{Intrinsic Motivation}

Random Network Distillation (RND)~\citep{burda2019exploration} is a widely-used intrinsic motivation method that provides exploration bonuses based on prediction error between a fixed random network and a trainable predictor. The intrinsic reward is high for novel states and decreases as the predictor learns. RND is domain-agnostic and requires no task-specific knowledge, making it a natural comparison baseline for our LLM-guided reward design approach.

\section{Method}
\label{sec:method}

\subsection{Overview}

\begin{figure}[t]
  \centering
  \resizebox{\textwidth}{!}{%
  \begin{tikzpicture}[
    node distance=1.2cm and 1.4cm,
    box/.style={draw, rounded corners, minimum height=0.9cm, minimum width=2.0cm,
                align=center, font=\small},
    data/.style={box, fill=blue!10},
    proc/.style={box, fill=orange!15},
    result/.style={box, fill=green!12},
    arrow/.style={-{Stealth[length=5pt]}, thick},
    label/.style={font=\scriptsize, midway},
  ]
    \node[data] (env) {Environment\\Description};
    \node[proc, right=of env] (llm) {\textbf{LLM}\\(Claude / GPT)};
    \node[data, right=of llm] (fn) {\texttt{reward\_fn}\\Python code};
    \node[proc, right=of fn] (ppo) {\textbf{PPO}\\Training};
    \node[result, right=of ppo] (agent) {Trained\\Agent};

    \draw[arrow] (env) -- (llm) node[label, above] {prompt};
    \draw[arrow] (llm) -- (fn) node[label, above] {generate};
    \draw[arrow] (fn) -- (ppo) node[label, above] {shape};
    \draw[arrow] (ppo) -- (agent) node[label, above] {3000 eps};

    \node[proc, below=1.4cm of ppo] (probe) {\textbf{Probe}\\(500 eps)};
    \node[proc, below=1.4cm of fn] (diag) {\textbf{Auto-}\\Diagnostics};

    \draw[arrow] (fn) -- (probe) node[label, right] {short run};
    \draw[arrow] (probe) -- (diag) node[label, above] {metrics};
    \draw[arrow, color=red!70!black] (diag) -- (llm)
        node[label, left, color=red!70!black, text width=1.8cm, align=center]
        {feedback +\\prev.\ code};

    \begin{scope}[on background layer]
      \draw[draw=purple!60, dashed, thick, rounded corners]
        ($(diag.south west)+(-0.3,-0.35)$) rectangle
        ($(probe.north east)+(0.3,0.35)$);
      \node[font=\scriptsize, purple!70] at
        ($(diag.south)!0.5!(probe.south)+(0,-0.55)$)
        {Iterative Refinement Loop ($\leq$3 rounds)};
    \end{scope}

    \node[box, fill=yellow!15, below=1.4cm of llm, minimum width=1.8cm] (valid) {Sandbox\\Validation};
    \draw[arrow] (llm) -- (valid) node[label, right] {};
    \draw[arrow] (valid) -- (fn) node[label, below] {pass};

  \end{tikzpicture}%
  }
  \caption{Overview of LLM-Guided Reward Design. The LLM generates a reward-shaping function from a natural-language environment description, validated in a sandbox before use. In iterative mode (dashed), short probe runs provide diagnostics for up to 3 refinement rounds.}
  \label{fig:pipeline}
\end{figure}
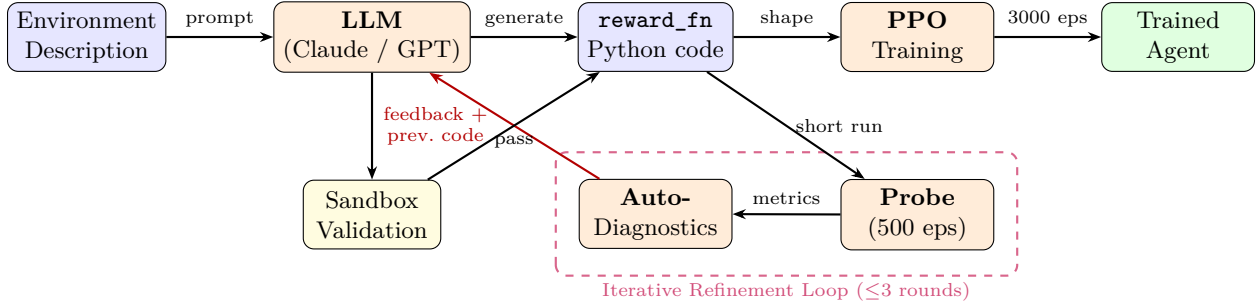

Our pipeline has three components (Figure~\ref{fig:pipeline}):
\begin{enumerate}
  \item \textbf{Reward Generation}: An LLM receives a natural-language description of the environment and produces a Python function \texttt{reward\_fn(obs, action, reward, terminated, truncated, info, state)} that returns a shaped reward and updated state dictionary.
  \item \textbf{RL Training}: A standard PPO~\citep{schulman2017ppo} agent trains using the shaped reward. The environment wrapper provides structured information (agent position, carrying status, event text) in the \texttt{info} dictionary.
  \item \textbf{Iterative Refinement} (optional): After a short probe training run, automated diagnostics assess whether the reward function exhibits failure modes. The diagnosis and training metrics are fed back to the LLM, which produces a revised function.
\end{enumerate}

\subsection{Reward Function Interface}

The generated reward function receives the full transition tuple plus a mutable \texttt{state} dictionary that persists across steps within an episode (reset between episodes). This enables one-time bonuses:

\begin{lstlisting}[style=python]
def reward_fn(obs, action, reward, terminated,
              truncated, info, state):
    shaped = reward
    if not state.get("key_picked_up"):
        if "picked up" in info["event_text"].lower():
            shaped += 0.2
            state["key_picked_up"] = True
    return shaped, state
\end{lstlisting}

The \texttt{info} dictionary exposes: \texttt{agent\_pos}, \texttt{carrying}, \texttt{event\_text} (natural-language event descriptions), \texttt{step\_count}, and \texttt{max\_steps}.

\subsection{LLM Prompt Design}

The broad evaluation uses a basic prompt containing the task description, a compact interface summary, and reward-design constraints; LLM Iterative additionally receives the previous reward source, probe metrics, and failure-mode feedback. The prompt-matched controls instead use an informed prompt that fully specifies all seven actions, the three observation channels, and every guaranteed \texttt{info} field. Informed one-shot and refinement receive identical static checks for reward accumulation, interface semantics, and weak shaping; refinement differs only by receiving the previous code, probe metrics, and dynamically triggered feedback. Appendix~\ref{app:prompts} gives the exact templates.

\subsection{Automated Diagnostics}
\label{sec:diagnostics}

After each probe training run, we compute three diagnostics (exact thresholds in Table~\ref{tab:thresholds}; supporting evidence in Section~\ref{sec:diagnostics_validation}):
\begin{itemize}
  \item \textbf{High-return/low-success mismatch}: If mean shaped reward $>0.5$ and task success rate $<0.2$, the rule flags possible accumulation of shaping reward without task completion. This is an operational heuristic, not proof of reward hacking. It is disabled for dense-reward locomotion tasks, which lack a binary success signal and use return-trend analysis instead (Section~\ref{sec:locomotion_failure}).
  \item \textbf{Possible shaping weakness}: If both success rate $< 0.1$ and mean reward $< 0.1$, the rule flags shaping that may be too weak or inert to guide exploration.
  \item \textbf{Learning plateau}: If success rate has stagnated (improvement $< 5\%$ over the probe window) after $> 1{,}000$ probe episodes, the agent may be stuck in a local optimum. This gate is inactive under the default 500-episode MiniGrid probes and is retained only for longer-probe variants.
\end{itemize}

These diagnostics are formatted as natural-language warnings in the refinement prompt. The control experiments (Section~\ref{sec:diagnostics_validation}) show that taxonomy-guided feedback is useful but imperfect: reward-hacking detection can generate false positives on some environments, shaping-weakness feedback is useful on moderately difficult tasks, and the plateau rule requires longer probes before it can be interpreted causally.

\subsection{Iterative Refinement Algorithm}

Algorithm~\ref{alg:iterative} formalizes the iterative refinement procedure.

\begin{center}
\refstepcounter{algorithm}\label{alg:iterative}
\textbf{Algorithm~\thealgorithm: LLM-guided iterative reward design}
\vspace{0.5ex}

\small
\begin{tabular}{@{}rl@{}}
\toprule
1 & $f \leftarrow \text{LLM}(\mathcal{D})$; validate $f$ in the sandbox. \\
2 & \textbf{for} $k = 0,\ldots,K-1$ \textbf{do} \\
3 & \quad $\pi_k \leftarrow \text{PPO}(f, N_p)$ \hfill probe training \\
4 & \quad $m_k \leftarrow \text{Evaluate}(\pi_k)$ \hfill success rate and reward \\
5 & \quad \textbf{if} $m_k.\text{success\_rate} \geq \tau$ \textbf{then break}. \\
6 & \quad $d_k \leftarrow \text{Diagnose}(m_k)$ \hfill failure-mode feedback \\
7 & \quad $f \leftarrow \text{LLM}(\mathcal{D}, f, m_k, d_k)$; validate $f$. \\
8 & $\pi^* \leftarrow \text{PPO}(f, N_f)$ \hfill full training \\
\bottomrule
\end{tabular}
\end{center}

\subsection{Safety and Validation}

Generated reward functions are executed in a sandboxed namespace with restricted \texttt{\_\_builtins\_\_} (only \texttt{math}, \texttt{abs}, \texttt{min}, \texttt{max}, \texttt{len}, etc.). Functions that fail validation (syntax errors, runtime errors on dummy inputs) trigger up to 3 retries with error feedback.

\subsection{Extension to Continuous Control}

For MuJoCo environments, we adapt the pipeline as follows: (1)~the \texttt{info} dictionary exposes position-based fields (\texttt{distance\_to\_target}, \texttt{end\_effector\_pos}, \texttt{target\_pos}) and event text (``reached target'', ``moved closer''), (2)~the LLM prompt is adapted for continuous observations (flat vectors) and continuous actions, emphasizing distance-based progress bonuses over event-driven bonuses, (3)~the PPO trainer uses an MLP policy with Gaussian actor, running observation normalization, and MuJoCo-appropriate hyperparameters (rollout length 2048, 10 PPO epochs, no entropy bonus). The reward function interface remains identical, ensuring the same iterative refinement protocol applies.

For locomotion environments (HalfCheetah, Hopper), the LLM designs an alternative reward function accessing velocity, position, and control cost from the \texttt{info} dictionary. The ``no shaping'' baseline uses the default Gymnasium reward. Since locomotion tasks have no binary success criterion, we evaluate using mean episode return over the final 100 episodes.

\section{Interpretation}
\label{sec:interpretation}

We use reward-shaping theory to distinguish our heuristic rewards from shaping methods with policy-invariance guarantees.

\subsection{Relation to Potential-Based Shaping}

Potential-based shaping preserves the optimal policy when the added reward has the form $F(s,s')=\gamma\Phi(s')-\Phi(s)$~\citep{ng1999policy}. Our generated rewards are not constrained to this form: they do not enforce the discount correction and may include components such as stagnation penalties. They therefore carry no policy-invariance guarantee, and we do not interpret them as potential-based shaping.

Potential-based shaping is useful here only as a theoretical contrast. A one-time milestone bonus avoids repeated accumulation, but it can still change trajectory preferences. Repeated per-step bonuses are more directly problematic in finite-horizon or terminating tasks because they can favor longer trajectories or delayed termination. Dynamic-potential results likewise require explicit structure rather than arbitrary time-varying bonuses~\citep{devlin2012dynamic}.

\subsection{Iterative Refinement Through a Variance-Reduction Lens}

Consider the space of reward functions $\mathcal{F}$ that an LLM can generate. One-shot generation samples $f \sim P_{\text{LLM}}(\mathcal{F} | \mathcal{D})$ conditioned on the environment description $\mathcal{D}$. The resulting policy performance $J(f) = \mathbb{E}_{\pi_f}[R]$ is a random variable with variance:
\begin{equation}
\text{Var}[J] = \text{Var}_f[\mathbb{E}_\pi[R | f]] + \mathbb{E}_f[\text{Var}_\pi[R | f]]
\end{equation}
The first term is LLM generation variance; the second is RL training variance. This equation describes the ideal marginal decomposition. Our anchored estimates in Section~\ref{sec:variance_decomp} use single-seed conditioning as a directional approximation, while the fully crossed $5 \times 5$ design in Table~\ref{tab:crossed_variance} realizes the random-effects partition directly on a representative subset.

Iterative refinement can be viewed as a conditional sampling procedure: rather than drawing independent samples from $P_{\text{LLM}}(\mathcal{F} | \mathcal{D})$, each refinement step conditions on the previous function and its training diagnostics, sampling from $P_{\text{LLM}}(\mathcal{F} | \mathcal{D}, f_k, d_k)$. This can be interpreted as reducing effective variability in the sampled reward functions by shifting generation toward functions that avoid diagnosed failure modes. The approach should be most useful when LLM generation variance dominates, which is consistent with the pattern we observe for discrete MiniGrid tasks, and should provide limited benefit when RL variance dominates, as in the continuous MuJoCo stress tests.

\section{Experimental Setup}
\label{sec:experiments}

\subsection{Environments}

We evaluate on nine MiniGrid environments and four MuJoCo environments (Table~\ref{tab:envs}):

\begin{table}[H]
\centering
\footnotesize
\setlength{\tabcolsep}{2pt}
\begin{tabular}{@{}lllc@{}}
\toprule
\textbf{Environment} & \textbf{Type} & \textbf{Key Challenge} & \textbf{Sub.} \\
\midrule
LavaGap-S5 & Grid & Navigation + avoidance & 2 \\
DoorKey-5$\times$5 & Grid & Key $\to$ door $\to$ goal & 3 \\
DoorKey-8$\times$8 & Grid & Larger search space & 3 \\
UnlockPickup & Grid & Key $\to$ door $\to$ box & 4 \\
KeyCorridor-S3R1 & Grid & Corridor navigation & 3 \\
BlockedUnlockPickup & Grid & Move ball, key, door, box & 4 \\
FourRooms & Grid & Multi-room navigation & 1 \\
MultiRoom-N6 & Grid & Long-horizon (6 rooms) & 1 \\
ObstructedMaze-2Dlh & Grid & Hidden key, obstruction & 4 \\
\midrule
Reacher-v4 & Cont. & 2-DOF arm $\to$ target & 1 \\
FetchReach-v4 & Cont. & 7-DOF gripper $\to$ goal & 1 \\
HalfCheetah-v4 & Cont. & Forward locomotion & 1 \\
Hopper-v4 & Cont. & Balance + locomotion & 1 \\
\bottomrule
\end{tabular}
\caption{Evaluation environments. Grid = MiniGrid discrete, Cont.\ = MuJoCo continuous. Sub.\ = sequential subgoals. The three new environments (FourRooms, MultiRoom-N6, ObstructedMaze) test longer planning horizons and harder exploration. Locomotion tasks feature dense default rewards and no binary success criterion.}
\label{tab:envs}
\end{table}

\subsection{Conditions}

We compare five conditions:
\begin{enumerate}
  \item \textbf{No Shaping}: Baseline PPO with only the sparse environment reward.
  \item \textbf{LLM One-Shot}: A single LLM call generates the reward function.
  \item \textbf{LLM Iterative}: Up to 3 refinement iterations with probe runs (500 episodes for MiniGrid, 200 for MuJoCo). Dense-reward locomotion uses the return-aware diagnostic adaptation in the main results; the original success-diagnostic variant is reported separately as a broken baseline.
  \item \textbf{Hand-Crafted}: Human-designed reward functions with tuned bonus magnitudes.
  \item \textbf{RND}: Random Network Distillation~\citep{burda2019exploration} intrinsic motivation as a domain-agnostic exploration baseline. The main table uses a fixed, untuned coefficient of $0.1$ across environments; predictor architecture: 3-layer MLP (256--256--128). We separately report a post-hoc coefficient sweep on the two DoorKey tasks.
\end{enumerate}

The 13-environment evaluation compares these end-to-end protocols. Because LLM One-Shot and LLM Iterative differ in both static guidance and access to probe feedback, their performance gap does not isolate the incremental effect of refinement. The prompt-matched evaluation in Section~\ref{sec:backbone_matched} holds the API description and static reward-design guidance fixed.

\subsection{Training Details}

All conditions use PPO~\citep{schulman2017ppo}. The main one-shot, no-shaping, hand-crafted, RND, and MiniGrid iterative conditions are evaluated across \textbf{10 random seeds} (42, 123, 456, 789, 1024, 2048, 3141, 4096, 5555, 7777). Exceptions are stated where they occur: return-aware locomotion refinement uses 6 completed seeds, and some secondary controls have fewer successful runs due to reward-function validation failures.

\textbf{MiniGrid hyperparameters}: learning rate $3 \times 10^{-4}$, $\gamma = 0.99$, GAE $\lambda = 0.95$, clip ratio 0.2, entropy coefficient 0.1, 2 PPO epochs, batch size 64, rollout length 512. Observations are normalized by dividing by 10 (MiniGrid symbolic values range 0--10). Training budget: 3{,}000 episodes.

\textbf{MuJoCo hyperparameters}: learning rate $3 \times 10^{-4}$, $\gamma = 0.99$, GAE $\lambda = 0.95$, clip ratio 0.2, entropy coefficient 0.0, 10 PPO epochs, batch size 64, rollout length 2{,}048. Observations use running mean/std normalization. MLP policy with 256-unit hidden layers and Gaussian actor (learnable log-std) with tanh squashing. Training budgets: 1{,}000 episodes for Reacher, HalfCheetah, and Hopper; 3{,}000 episodes for FetchReach (to account for its 7-DOF complexity).

The LLM used for reward generation is Claude Haiku (\texttt{claude-haiku-4-5-20251001}) with temperature 0.4.

For the prompt-matched backbone control, we additionally evaluate Claude Opus~5 (\texttt{claude-opus-5}, high-effort setting) and DeepSeek V4 Pro (\texttt{deepseek-v4-pro}, thinking enabled), alongside Haiku. DeepSeek V4 Pro has released weights~\citep{deepseek2026v4}; our runs use its hosted API endpoint and save the returned model text and accepted reward code. Each backbone is tested on DoorKey-8$\times$8 and KeyCorridor with the same 10 seeds.

\textbf{Reproducibility artifacts.} The supplementary release includes prompts, generated reward functions, thresholds, seed-level logs, configurations, scripts, returned model text, and validator outcomes. The frozen reward functions allow the RL runs to be repeated without calling the hosted models again.

\textbf{Episode budget.} The MiniGrid iterative condition uses up to $K \times N_p + N_f = 3 \times 500 + 3{,}000 = 4{,}500$ total training episodes (50\% overhead). The MuJoCo iterative condition uses up to $3 \times 200 + N_f$ episodes, where $N_f$ is the environment-specific full-training budget.

\section{Results}
\label{sec:results}

\subsection{Main Results}

Table~\ref{tab:main_results} presents final performance for each condition, computed over the last 100 training episodes and averaged across random seeds.

\begin{table}[H]
\centering
\footnotesize
\setlength{\tabcolsep}{3pt}
\begin{tabular}{@{}llccccc@{}}
\toprule
\textbf{Environment} & \textbf{Metric} & \textbf{No Shaping} & \textbf{LLM One-Shot} & \textbf{LLM Iterative} & \textbf{Hand-Crafted} & \textbf{RND} \\
\midrule
\multicolumn{7}{l}{\textit{MiniGrid (discrete, 3{,}000 episodes)}} \\
LavaGap-S5          & SR & $\mathbf{64.4 \pm 23.7\%}$ & $38.2 \pm 19.3\%$ & $54.4 \pm 13.7\%$ & $63.1 \pm 15.6\%$ & $52.3 \pm 18.7\%$ \\
DoorKey-5$\times$5  & SR & $99.8 \pm 0.4\%$  & $85.3 \pm 31.4\%$ & $99.7 \pm 0.5\%$  & $\mathbf{99.9 \pm 0.3\%}$ & $13.2 \pm 6.6\%$ \\
DoorKey-8$\times$8  & SR & $2.3 \pm 1.1\%$   & $58.3 \pm 47.3\%$ & $97.6 \pm 3.0\%$           & $\mathbf{100.0 \pm 0.0\%}$ & $0.1 \pm 0.3\%$ \\
UnlockPickup        & SR & $0.1 \pm 0.3\%$   & $15.8 \pm 7.8\%$  & $\mathbf{24.3 \pm 10.3\%}$ & $18.9 \pm 5.2\%$ & $0.0 \pm 0.0\%$ \\
KeyCorridor         & SR & $11.7 \pm 5.3\%$  & $31.2 \pm 37.1\%$  & $\mathbf{86.7 \pm 28.5\%}$ & $29.2 \pm 17.8\%$ & $10.4 \pm 3.6\%$ \\
BlockedUnlockPickup & SR & $0.0 \pm 0.0\%$   & $0.7 \pm 0.8\%$   & $0.6 \pm 1.0\%$   & $\mathbf{0.8 \pm 0.8\%}$ & $0.0 \pm 0.0\%$ \\
FourRooms           & SR & $\mathbf{27.2 \pm 4.2\%}$ & $19.9 \pm 6.5\%$ & $23.3 \pm 7.2\%$ & $17.7 \pm 2.9\%$ & $6.1 \pm 1.2\%$ \\
MultiRoom-N6        & SR & $0.0 \pm 0.0\%$ & $\mathbf{0.5 \pm 1.0\%}$ & $0.3 \pm 0.7\%$ & $0.0 \pm 0.0\%$ & $0.0 \pm 0.0\%$ \\
ObstructedMaze      & SR & $0.5 \pm 0.7\%$ & $\mathbf{7.4 \pm 5.0\%}$ & $6.6 \pm 2.3\%$ & $6.5 \pm 3.1\%$ & $0.0 \pm 0.0\%$ \\
\midrule
\multicolumn{7}{l}{\textit{MuJoCo Reaching (continuous, 1{,}000 / 3{,}000 episodes)}} \\
Reacher-v4          & SR & $65.0 \pm 7.6\%$  & $\mathbf{77.0 \pm 4.1\%}$ & $73.7 \pm 6.5\%$ & $75.4 \pm 6.6\%$ & $67.5 \pm 8.6\%$ \\
FetchReach-v4       & SR & $6.5 \pm 2.4\%$    & $7.5 \pm 2.6\%$    & $\mathbf{12.8 \pm 8.0\%}$ & $8.0 \pm 3.4\%$    & $6.5 \pm 2.2\%$ \\
\midrule
\multicolumn{7}{l}{\textit{MuJoCo Locomotion (continuous, 1{,}000 episodes; iterative = return-aware, $n=6$)}} \\
HalfCheetah-v4      & Ret & $1864 \pm 749$     & $2190 \pm 405$  & $2215 \pm 485$   & $2287 \pm 497$   & $1580 \pm 641$ \\
Hopper-v4           & Ret & $453 \pm 104^{\dagger}$ & $485 \pm 101$   & $472 \pm 112$     & $462 \pm 163$    & $549 \pm 178$ \\
\bottomrule
\end{tabular}
\caption{Main results across 13 environments (last 100 episodes, mean $\pm$ sample std). SR = success rate; Ret = mean episode return. \textbf{Bold} indicates best performance per environment for MiniGrid and reaching tasks; no locomotion cell is bolded because those rows are boundary diagnostics with no robust corrected improvement. The One-Shot and Iterative columns are end-to-end protocols with different static guidance; Table~\ref{tab:backbone_matched} provides the prompt-matched comparison. We report 12 designated comparisons in Appendix~\ref{app:primary_stats}. The iterative column reports standard diagnostic refinement for MiniGrid and reaching tasks, and return-aware refinement for dense-reward locomotion. Return-aware locomotion results use the first 6 completed canonical seeds; other main results use 10 seeds. FetchReach trained for 3{,}000 episodes; all other MuJoCo environments for 1{,}000. $^\dagger$ The Hopper no-shaping value is the 1{,}000-episode main-budget result; the budget-matched 1{,}600-episode no-shaping result is $882$ in Section~\ref{sec:budget_control}. RND uses a fixed, untuned coefficient $c=0.1$ in this table; Table~\ref{tab:rnd_sweep} reports a coefficient sweep for the two DoorKey rows.}
\label{tab:main_results}
\end{table}

We do not interpret these results as showing that LLM rewards generally dominate hand-crafted rewards, intrinsic motivation, or continuous-control baselines. The One-Shot and Iterative columns compare end-to-end protocols, but their gap does not isolate corrective feedback because the protocols also differ in static guidance. Sections~\ref{sec:metrics_only} and~\ref{sec:backbone_matched} provide the corresponding prompt controls.

Key findings:

\begin{itemize}
  \setlength{\topsep}{4pt}
  \setlength{\partopsep}{0pt}
  \setlength{\itemsep}{2pt}
  \setlength{\parsep}{0pt}
  \setlength{\parskip}{0pt}
  \item \textbf{One-shot LLM design is unreliable}: While it improves over no shaping on DoorKey-8$\times$8 ($2.3\% \to 58.3\%$), it shows high variance (sample std up to $47.3\%$) and underperforms no shaping on LavaGap ($38.2\%$ vs.\ $64.4\%$). On DoorKey-5$\times$5, one-shot drops to $85.3\%$ while all other non-RND conditions reach $\sim$100\%.

  \item \textbf{Iterative refinement provides its clearest gains on sparse structured tasks}: KeyCorridor improves from $31.2\%$ one-shot to $86.7\%$ iterative, and DoorKey-8$\times$8 reaches $97.6\%$ (vs.\ $58.3\%$ one-shot). These are end-to-end protocol differences; the prompt-matched controls in Section~\ref{sec:backbone_matched} provide the stricter test of corrective refinement.

  \item \textbf{Fixed-coefficient RND is mixed and coefficient-sensitive}: With the untuned coefficient used in Table~\ref{tab:main_results}, RND hurts performance on both DoorKey tasks. A coefficient sweep shows that $c=0.01$ solves DoorKey-5$\times$5, while none of the three tested coefficients solves DoorKey-8$\times$8. We therefore treat RND as a contextual intrinsic-motivation baseline, not as a tuned baseline claim.

  \item \textbf{The strongest positive results occur where one-shot generation is unreliable but subgoals are semantically exposed}: DoorKey-8$\times$8 reaches $97.6\%$ and KeyCorridor reaches $86.7\%$ iterative success. The KeyCorridor mean is high but unstable (sample std $28.5\%$), and later stress tests show that diagnostic thresholds and \texttt{event\_text} signals can substantially change outcomes. On LavaGap, iterative refinement reaches $54.4\%$, below no shaping ($64.4\%$), showing that the pipeline does not help uniformly. On BlockedUnlockPickup, all conditions achieve $\leq 0.8\%$, indicating a task beyond the reach of reward shaping alone.

  \item \textbf{Richer semantic interfaces are not always better}: The headline KeyCorridor result ($86.7\%$, $n=10$) is essentially matched by a 5-seed rerun that removes \texttt{event\_text} from both the prompt and runtime interface ($86.8\%$), while the paired with-events rerun reaches only $55.2\%$. The ablation uses independent LLM generations, so the near-equality should be read as a scope warning rather than a paired replication.

  \item \textbf{Harder environments reveal limits}: On FourRooms, no shaping ($27.2\%$) outperforms all reward-shaped conditions. On ObstructedMaze, LLM one-shot ($7.4\%$) and iterative ($6.6\%$) improve over no shaping ($0.5\%$). MultiRoom-N6 remains near 0\% across all conditions.
\end{itemize}

\begin{table}[ht]
\centering
\small
\begin{tabular}{@{}lccc@{}}
\toprule
\textbf{Environment} & \textbf{RND $c=0.01$} & \textbf{RND $c=0.1$} & \textbf{RND $c=0.5$} \\
\midrule
DoorKey-5$\times$5 & $\mathbf{100.0 \pm 0.0\%}$ & $14.8 \pm 5.6\%$ & $4.5 \pm 4.5\%$ \\
DoorKey-8$\times$8 & $\mathbf{0.9 \pm 0.9\%}$ & $0.0 \pm 0.0\%$ & $0.4 \pm 0.8\%$ \\
\bottomrule
\end{tabular}
\caption{RND coefficient sweep on the two DoorKey tasks (10 seeds, 3{,}000 episodes, final-100 success rate). The fixed $c=0.1$ baseline in Table~\ref{tab:main_results} was badly mistuned on DoorKey-5$\times$5, where $c=0.01$ solves the task, but tuning over this small grid does not rescue RND on DoorKey-8$\times$8.}
\label{tab:rnd_sweep}
\end{table}

\subsection{Learning Curves}

\begin{figure}[H]
  \centering
  \includegraphics[width=0.90\columnwidth]{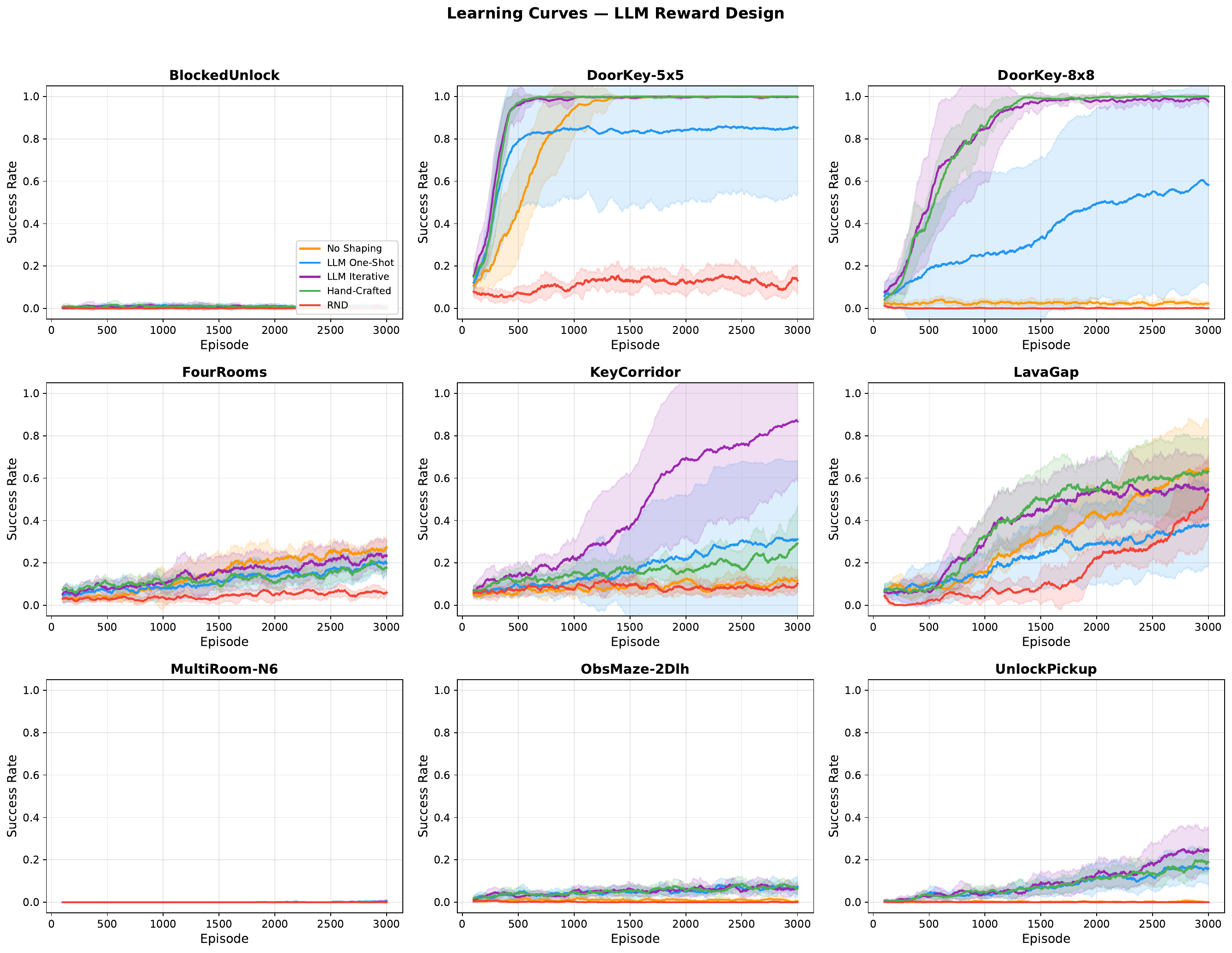}
  \caption{Learning curves (success rate vs.\ episode, smoothed over 100 episodes). Shaded regions show $\pm$1 sample std across seeds.}
  \label{fig:learning_curves}
\end{figure}

Figure~\ref{fig:learning_curves} shows learning curves across the core sparse tasks. On DoorKey-8$\times$8, the no-shaping baseline reaches $2.3\%$ success, while iterative refinement reaches $97.6\%$. LavaGap shows the opposite pattern: iterative refinement reaches $54.4\%$, below no shaping ($64.4\%$) and hand-crafted shaping ($63.1\%$).

\subsection{Post-Refinement Training Episodes}

\begin{figure}[H]
  \centering
  \includegraphics[width=\columnwidth]{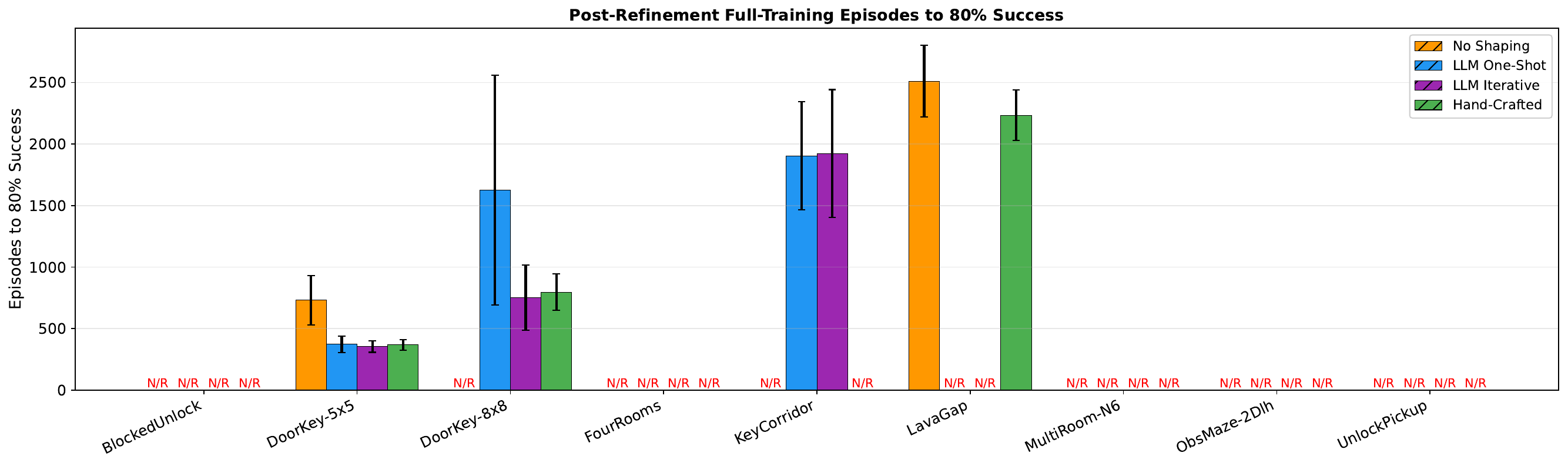}
  \caption{Full-training episodes to reach 80\% success after reward selection/refinement. Probe episodes are excluded here; total-budget controls are reported in Section~\ref{sec:budget_control}. Hatched bars (N/R) indicate the threshold was never reached within 3{,}000 full-training episodes.}
  \label{fig:sample_efficiency}
\end{figure}

Figure~\ref{fig:sample_efficiency} compares the number of full-training episodes required to reach 80\% success rate after the reward function has been selected or refined. Probe episodes used during reward design are excluded from this figure; total environment-interaction controls are reported separately in Section~\ref{sec:budget_control}. Under this post-refinement training view, the iterative condition reaches the threshold faster than the hand-crafted baselines on DoorKey-5$\times$5 and DoorKey-8$\times$8.

\section{Failure Taxonomy}
\label{sec:failure_taxonomy}

The main results show that one-shot LLM reward generation is useful but unreliable. To make the failure structure actionable, we classify generated reward functions into two recurring failure modes and one rarer weak-shaping case (Table~\ref{tab:failure_modes}).

\begin{table}[H]
\centering
\footnotesize
\begin{tabular}{@{}p{2.1cm}p{4.15cm}@{}}
\toprule
\textbf{Failure Mode} & \textbf{Description} \\
\midrule
Reward Flooding & Continuous per-step rewards (e.g., $+0.02$ for every forward action) dominate the shaped reward, masking task-completion signal. \\
\midrule
Semantic/API Misunderstanding & LLM misuses environment semantics or APIs (e.g., maps MiniGrid action indices incorrectly, or accesses undocumented \texttt{info} fields). \\
\midrule
Weak / Negligible Shaping & Bonus magnitudes too small relative to the sparse reward, providing insufficient gradient signal; rare in our audit. \\
\bottomrule
\end{tabular}
\caption{Failure modes observed under the basic one-shot prompt. Reward flooding and semantic/API misunderstanding recur in our qualitative audit; weak or negligible shaping is observed more rarely and is harder to distinguish from optimization failure.}
\label{tab:failure_modes}
\end{table}

\textbf{Reward flooding} is one of the most damaging observed failure modes. In DoorKey-8$\times$8, one generated function assigned $+0.02$ per forward step, leading to high accumulated reward but only $\sim$5\% success rate---the agent learned to walk in circles collecting step rewards rather than solving the task. The iterative diagnostic detects this pattern (high reward, low success) and explicitly warns the LLM in the refinement prompt.

Semantic/API errors depend strongly on interface specification. Under the API-complete prompt used in the matched evaluation, all initial and final reward functions pass syntax and API lint, with no undocumented \texttt{info} fields or invalid action literals. The observed API failures therefore characterize the basic-prompt regime rather than an unavoidable property of LLM reward design.

The taxonomy defines the feedback vocabulary used in iterative refinement. The next section evaluates the contribution and limitations of that feedback.

\section{Diagnostics Evidence and Controls}
\label{sec:diagnostics_validation}

Aggregate performance alone does not establish that the diagnostic signals are reliable. We therefore treat the diagnostics as empirically useful but imperfect heuristics and evaluate them through targeted controls: a component-removal stress test (which detector families are risky?), prompt controls against metrics-only and static-vocabulary baselines (does taxonomy-rich feedback add value, and how much is due to the static taxonomy vocabulary?), a qualitative audit of recurring failure patterns, and sensitivity checks for threshold and probe design. Table~\ref{tab:diagnostics_evidence} summarizes what each control supports and what it does not establish. Appendix~\ref{app:thresholds} lists all trigger values used in the experiments.

Unless stated otherwise, control experiments in this section are independent lower-seed reruns, not exact reproductions of the 10-seed means in Table~\ref{tab:main_results}; captions report the seed counts and protocol differences.

\begin{table}[H]
\centering
\scriptsize
\setlength{\tabcolsep}{2pt}
\begin{tabular}{@{}p{2.35cm}p{3.15cm}p{4.25cm}p{4.15cm}@{}}
\toprule
\textbf{Control} & \textbf{Question answered} & \textbf{Main result} & \textbf{Limitation} \\
\midrule
Metrics-only re-prompting & Does taxonomy-rich feedback add value beyond retrying with raw metrics? & Removing diagnostic tags and static failure-mode vocabulary causes large hard-task drops (DoorKey-8$\times$8: $97.6\%\!\to\!68.6\%$; KeyCorridor: $86.7\%\!\to\!11.5\%$). & Prompt-level control; intentionally removes both static and dynamic taxonomy information. \\
\midrule
Static-vocabulary control & Are dynamic per-iteration labels isolated as the mechanism? & Keeping static failure-mode vocabulary but removing dynamic labels recovers much of the gap (DoorKey-8$\times$8: $87.6\%$; KeyCorridor: $70.7\%$). & Full diagnostics still trend higher, but full-vs-static gaps are not statistically resolved at $n=10$. \\
\midrule
Component-removal stress test & Which detector families create sensitivity? & Shaping-weakness removal hurts DoorKey; reward-hacking removal exposes false positives on KeyCorridor. & Low-seed independent reruns; the plateau gate is inactive under 500-episode probes. \\
\midrule
Threshold sensitivity & Are fixed trigger values brittle? & Reward-hacking and shaping thresholds reveal calibration sensitivity. & No universal threshold; plateau rows are inactive under default probe length. \\
\midrule
Probe-length sensitivity & How much probe budget is needed? & Diagnostic quality depends on the observation window; longer probes reduce failures on hard tasks. & Weakens any claim that very short probes always suffice. \\
\midrule
Qualitative taxonomy audit & Which failure patterns recur? & Reward flooding and semantic/API misuse recur; weak shaping is rarer and harder to distinguish. & Descriptive evidence only; online trigger rules are evaluated separately through prompt controls and sensitivity analyses. \\
\midrule
Best-of-3 selection & Is improvement just selection among multiple samples? & Selection helps strongly; iterative has a higher KeyCorridor mean but the $n=10$ gap is not statistically resolved. & MuJoCo Best-of-3 has incomplete seeds due to validation failures. \\
\midrule
Episode-budget control & Is improvement just extra training time? & MiniGrid gains persist beyond extra training time; Hopper exposes a training-duration confound. & Does not remove all optimizer/reward interaction effects. \\
\bottomrule
\end{tabular}
\caption{Evidence matrix for the diagnostic mechanism. Environment, seed, and budget details are reported in the corresponding subsections. The controls support the debugging interpretation through converging evidence while also identifying calibration limits and boundary cases.}
\label{tab:diagnostics_evidence}
\end{table}

\subsection{Component-Removal Stress Test}
\label{sec:diagnostic_ablation}

We remove each diagnostic component one at a time---reward-hacking detection~(RH), shaping-weakness detection~(SW), and learning-plateau detection~(LP)---and compare against the full system on four representative environments. These are independent 3-seed reruns and provide a stress test for detector sensitivity rather than a causal decomposition of the 10-seed main evaluation.

\begin{table}[ht]
\centering
\small
\setlength{\tabcolsep}{4pt}
\begin{tabular}{@{}lcccc@{}}
\toprule
\textbf{Environment} & \textbf{Full} & \textbf{$-$RH} & \textbf{$-$SW} & \textbf{$-$LP$^\dagger$} \\
\midrule
DoorKey-5$\times$5  & $1.00$ & $1.00$         & $0.99$         & $1.00$  \\
DoorKey-8$\times$8  & $0.93$ & $0.95$         & $\mathbf{0.64}$ & $0.97$  \\
KeyCorridor         & $0.71$ & $1.00$         & $1.00$          & $0.18$ \\
UnlockPickup        & $0.21$ & $0.25$         & $0.23$          & $0.23$  \\
\bottomrule
\end{tabular}
\caption{Component-removal stress test: mean success rate from independent 3-seed reruns with one detector removed at a time. \textbf{Bold} marks substantial degradation ($>$15 points) outside the LP column. $^\dagger$Because the plateau gate requires $>1{,}000$ probe episodes, the $-$LP column reports an observed rerun drop rather than causal evidence about plateau warnings under the default 500-episode probes.}
\label{tab:ablation}
\end{table}

The stress test reveals an asymmetric sensitivity pattern. \textbf{Shaping-weakness detection matters on moderately difficult tasks}: removing it drops DoorKey-8$\times$8 from $93\%$ to $64\%$ ($\Delta = -29$ points), where the reward function must provide sufficient exploration guidance. The $-$LP KeyCorridor drop ($71\% \to 18\%$) is not interpreted as a clean plateau-detector effect: local-log inspection shows that the default 500-episode probes do not satisfy the $>1{,}000$-episode plateau eligibility gate, so plateau warnings do not fire in these runs. We therefore treat this row as evidence of lower-seed rerun sensitivity, not as proof that the plateau detector explains the KeyCorridor gains.

An unexpected finding is that \textbf{removing the reward-hacking detector improves performance} on KeyCorridor ($71\% \to 100\%$, $\Delta = +28$) and does not hurt DoorKey-8$\times$8 ($93\% \to 95\%$). This suggests the reward-hacking detector generates false positives on these environments, triggering unnecessary refinements that degrade the reward function. We retain the detector in the headline pipeline because it provides a modest UnlockPickup safeguard and catches pathological flooding cases, but practitioners should disable it or raise its cutoff when probe traces show useful progress despite low early success. These results suggest that shaping-weakness and reward-hacking diagnostics are useful but require calibration, while plateau detection should only be analyzed in protocols where its eligibility gate can actually fire.

\subsection{Taxonomy Prompt Controls}
\label{sec:metrics_only}

A natural question is whether the gains come from taxonomy-rich diagnostic feedback or merely from giving the LLM another attempt. We first compare the full diagnostic pipeline against a \emph{metrics-only baseline} that provides the LLM with identical numerical metrics (success rate, mean reward) but strips all dynamic diagnostic tags, failure-mode names, and static ``common failure modes to avoid'' guidance. The LLM receives only: ``the results are suboptimal; write an improved reward function.'' This is a strong prompt-level ablation of taxonomy information.

\begin{table}[H]
\centering
\small
\begin{tabular}{@{}lcc@{}}
\toprule
\textbf{Environment} & \textbf{Full Diagnostic} & \textbf{Metrics-Only} \\
\midrule
DoorKey-5$\times$5  & $99.7\%$  & $92.8 \pm 21.4\%$ \\
DoorKey-8$\times$8  & $97.6\%$  & $68.6 \pm 38.9\%$ \\
UnlockPickup        & $24.3\%$  & $6.8 \pm 7.7\%$   \\
KeyCorridor         & $86.7\%$  & $11.5 \pm 10.1\%$  \\
\bottomrule
\end{tabular}
\caption{Full diagnostic-guided refinement vs.\ metrics-only re-prompting (mean $\pm$ sample std; metrics-only uses 9--10 successful seeds, with UnlockPickup at $n=9$). The full diagnostic column reuses the 10-seed Table~\ref{tab:main_results} means. The taxonomy-rich feedback condition provides 29--75 percentage point improvements on the hardest tasks, but this prompt-level ablation removes both dynamic diagnostic labels and static failure-mode vocabulary.}
\label{tab:metrics_only}
\end{table}

Removing taxonomy-rich feedback causes large performance drops on the hardest tasks: DoorKey-8$\times$8 drops from $97.6\%$ to $68.6\%$ ($\Delta=29.0$~points; Welch $p=0.043$; bootstrap $95\%$ CI $[7.9,53.5]$~points), KeyCorridor from $86.7\%$ to $11.5\%$ ($\Delta=75.2$~points; $p=6.7{\times}10^{-6}$; CI $[56.0,90.7]$), and UnlockPickup from $24.3\%$ to $6.8\%$ ($\Delta=17.5$~points; $p=6.1{\times}10^{-4}$; CI $[9.4,24.7]$). DoorKey-5$\times$5 is saturated and the smaller gap is not statistically resolved ($p=0.33$). The metrics-only baseline is comparable to one-shot generation on the hard sparse tasks, providing evidence that \textbf{taxonomy-rich feedback, not mere re-prompting, is important to iterative refinement's effectiveness}.

To separate static taxonomy vocabulary from dynamic trigger labels, we add a stricter control on DoorKey-8$\times$8 and KeyCorridor. This condition keeps the static \texttt{COMMON FAILURE MODES TO AVOID} block and the numerical probe metrics, but removes per-iteration trigger strings such as \texttt{REWARD HACKING DETECTED}, \texttt{SHAPING TOO WEAK}, and \texttt{PLATEAU DETECTED}. Iteration-0 reward functions are shared with the full diagnostic condition when available, so the manipulation begins at refinement.

\begin{table}[H]
\centering
\small
\begin{tabular}{@{}lccc@{}}
\toprule
\textbf{Environment} & \textbf{Full Diagnostic} & \textbf{Static Vocab Only} & \textbf{Metrics-Only} \\
\midrule
DoorKey-8$\times$8 & $97.6 \pm 3.0\%$ & $87.6 \pm 30.6\%$ & $68.6 \pm 38.9\%$ \\
KeyCorridor        & $86.7 \pm 28.5\%$ & $70.7 \pm 42.9\%$ & $11.5 \pm 10.1\%$ \\
\bottomrule
\end{tabular}
\caption{Static-vocabulary prompt control (10 seeds). Static-vocab-only keeps the failure-mode vocabulary and raw metrics while removing dynamic per-iteration diagnostic labels. It recovers much of the metrics-only gap, especially on KeyCorridor, so the mechanism should be framed as taxonomy-guided prompting plus diagnostics rather than dynamic trigger labels alone.}
\label{tab:static_vocab_control}
\end{table}

The static-vocabulary control changes the mechanism interpretation. On DoorKey-8$\times$8, full diagnostics exceed static-vocabulary prompting by $10.0$ points, but the interval includes small or negative gaps (bootstrap CI $[-1.8,30.3]$; Welch $p=0.33$; $d=0.46$). Static vocabulary exceeds metrics-only by $19.0$ points, but that comparison is also not resolved at $n=10$ (CI $[-10.6,47.9]$; $p=0.24$; $d=0.54$). On KeyCorridor, static vocabulary captures most of the improvement over metrics-only ($70.7\%$ vs.\ $11.5\%$; $\Delta=59.2$ points; CI $[32.1,83.5]$; $p=0.0017$; $d=1.90$), while full diagnostics trend higher than static vocabulary by $16.0$ points but remain statistically unresolved (CI $[-13.6,46.7]$; $p=0.34$; $d=0.44$). We therefore interpret the prompt controls as evidence that the failure taxonomy is a major active ingredient; the online diagnostic labels may add value, but these experiments do not isolate them as the sole driver. A seed-level inspection illustrates why we avoid dismissing dynamic labels entirely: on DoorKey-8$\times$8 seed 789, static-vocab-only reaches only $1\%$ final-100 success, whereas the full diagnostic run reaches $100\%$. This single catastrophic static-vocab seed explains much of the DoorKey static-vocab variance and is suggestive, not conclusive, evidence that dynamic labels can rescue hard cases.

\subsection{Threshold Sensitivity}
\label{sec:threshold_sensitivity}

We vary each diagnostic threshold independently while keeping all other components fixed, evaluating on DoorKey-8$\times$8 and KeyCorridor with up to 5~successful seeds per configuration. Table~\ref{tab:threshold_sensitivity} reports mean success rates.

\vfill
\newpage
\begin{table}[H]
\centering
\footnotesize
\setlength{\tabcolsep}{7pt}
\begin{tabular}{@{}llccc@{}}
\toprule
\textbf{Sweep} & \textbf{Env} & \textbf{Low} & \textbf{Default} & \textbf{High} \\
\midrule
\multicolumn{2}{l}{\textit{RH cutoff} (0.3 / 0.5 / 1.0)} \\
& DoorKey-8$\times$8  & $80.0 \pm 44.7$ & $94.2 \pm 6.8$  & $98.6 \pm 2.6$ \\
& KeyCorridor         & $74.0 \pm 39.7$ & $53.2 \pm 44.8$ & $48.2 \pm 47.5$ \\
\midrule
\multicolumn{2}{l}{\textit{Success-rate bound} (0.1 / 0.2 / 0.3)} \\
& DoorKey-8$\times$8  & $89.8 \pm 18.5$ & $94.8 \pm 6.4$  & $85.8 \pm 16.6$ \\
& KeyCorridor         & $47.0 \pm 48.5$ & $35.4 \pm 36.5$ & $32.2 \pm 38.7$ \\
\midrule
\multicolumn{2}{l}{\textit{Plateau cutoff} (0.02 / 0.05 / 0.1)} \\
& DoorKey-8$\times$8  & $80.4 \pm 43.3$ & $94.2 \pm 6.8$  & $66.3 \pm 56.6$ \\
& KeyCorridor         & $\mathbf{100.0 \pm 0.0}$ & $53.2 \pm 44.8$ & $68.8 \pm 45.5$ \\
\midrule
\multicolumn{2}{l}{\textit{SW cutoff} (0.05 / 0.1 / 0.3)} \\
& DoorKey-8$\times$8  & $98.2 \pm 2.4$  & $94.2 \pm 6.8$  & $98.6 \pm 2.1$ \\
& KeyCorridor         & $71.0 \pm 39.3$ & $53.2 \pm 44.8$ & $66.0 \pm 46.7$ \\
\bottomrule
\end{tabular}
\caption{Threshold sensitivity: final-100 success rate in percent (mean $\pm$ sample std; up to 5~seeds per cell) when varying one threshold at a time. The success-rate-bound rows share the initial Haiku function and RL seed across settings; the other rows are independent reruns. The plateau gate cannot fire under the default 500-episode probe, so its rows measure rerun variability rather than a plateau-threshold effect.}
\label{tab:threshold_sensitivity}
\end{table}

The shaped-return and shaping-weakness sweeps are non-monotonic and differ by task. Directly varying the success-rate bound is likewise task-dependent: $0.2$ has the highest DoorKey mean, whereas $0.1$ has the highest KeyCorridor mean. The low-seed intervals are wide, and the plateau rows are not causally interpretable because that gate is inactive. We therefore retain $0.2$ as the fixed setting used in the main experiments, without treating it as a universal boundary between reward hacking and acceptable shaping.

\subsection{Probe-Length Sensitivity}
\label{sec:probe_length}

The probe run is the diagnostic observation window: too short, and the diagnostics may lack signal; too long, and compute is wasted. We evaluate probe lengths of 300, 500, and 800 episodes on DoorKey-8$\times$8 and KeyCorridor (3~seeds each).

\begin{table}[H]
\centering
\footnotesize
\setlength{\tabcolsep}{7pt}
\begin{tabular}{@{}lccc@{}}
\toprule
\textbf{Environment} & \textbf{300 eps} & \textbf{500 eps} & \textbf{800 eps} \\
\midrule
DoorKey-8$\times$8  & $0.67 \pm 0.58$ & $1.00 \pm 0.01$ & $0.98 \pm 0.03$ \\
KeyCorridor         & $0.49 \pm 0.43$ & $0.40 \pm 0.52$ & $\mathbf{1.00 \pm 0.00}$ \\
\bottomrule
\end{tabular}
\caption{Probe-length sensitivity: mean success rate $\pm$ sample std from independent 3-seed reruns. Longer probes provide more reliable diagnostic observations, especially on hard environments.}
\label{tab:probe_length}
\end{table}

Contrary to the intuition that shorter probes suffice for quick diagnosis, longer probes substantially stabilize KeyCorridor, though 800 episodes is not uniformly best on every environment. On KeyCorridor, probe$=$300 achieves $49\%$ while probe$=$800 reaches $100\%$ across all seeds---the longer observation window gives the LLM less noisy refinement metrics and reduces premature or misleading updates.

On DoorKey-8$\times$8, the 300-episode probe produces a catastrophic failure on one seed (SR$=$0\%) while the other two seeds succeed, suggesting that 300~episodes is below the threshold needed for reliable reward-function evaluation on this task. The default of 500~episodes provides a reasonable balance, but practitioners working on hard tasks may benefit from longer probes.

These results reinforce the paper's central thesis: the diagnostic system's effectiveness depends on the quality of its observations, not just the quality of its heuristics.

\subsection{Interface Ablation}
\label{sec:scope_ablation}

The method depends on a curated reward-function interface (Section~\ref{sec:limitations}). To test how much the specific \texttt{event\_text} field contributes, we strip it from both the LLM prompt and the runtime \texttt{info} dictionary, forcing the LLM to rely on other semantic fields (\texttt{carrying}, \texttt{agent\_pos}, \texttt{step\_count}). We compare the standard pipeline (``with events'') against this reduced interface (``without events'') on DoorKey-8$\times$8 and KeyCorridor (5~seeds each, iterative refinement with 3~iterations). The two ablation conditions use the same five canonical seed IDs and independent LLM generations; those seed IDs are a subset of the 10-seed headline run.

\begin{table}[ht]
\centering
\footnotesize
\setlength{\tabcolsep}{3pt}
\begin{tabular}{@{}lccc@{}}
\toprule
\textbf{Env} & \textbf{With evt.} & \textbf{No evt.} & \textbf{No $-$ With} \\
\midrule
DoorKey-8$\times$8  & $96.6 \pm 7.1\%$  & $94.4 \pm 6.9\%$ & $-2.2$ \\
KeyCorridor         & $55.2 \pm 41.0\%$ & $86.8 \pm 29.5\%$ & $+31.6$ \\
\bottomrule
\end{tabular}
\caption{Interface ablation: success rate (\%) $\pm$ sample std (5~seeds) when \texttt{event\_text} is removed from the LLM prompt and runtime info dict. On KeyCorridor, removing \texttt{event\_text} \emph{improves} performance.}
\label{tab:scope_ablation}
\end{table}

The results are nuanced. On DoorKey-8$\times$8, removing \texttt{event\_text} costs only $2.2$ percentage points---within noise---because the LLM can infer subgoal completion from \texttt{carrying} transitions. On KeyCorridor, removing \texttt{event\_text} improves performance by $31.6$ points ($55.2\% \to 86.8\%$), essentially matching the 10-seed headline mean of $86.7\%$. Because the ablation uses independent LLM generations, this near-equality is not a paired replication of the headline result; it is nevertheless a warning against framing the KeyCorridor result as specifically event-text-driven. Inspection of the generated reward functions suggests that \texttt{event\_text} can cause the LLM to write over-specific reward logic, whereas no-events functions rely more heavily on position- and carrying-based heuristics.

This finding refines the paper's scope claim: the method's effectiveness depends on having \emph{some} structured semantic interface, but it does not critically depend on any single field. Rich event descriptions can help, hurt, or be neutral depending on how well they align with the task's subgoal structure.

\section{Analysis}
\label{sec:analysis}

\subsection{Refinement Trajectory}

\begin{figure}[ht]
  \centering
  \includegraphics[width=\columnwidth]{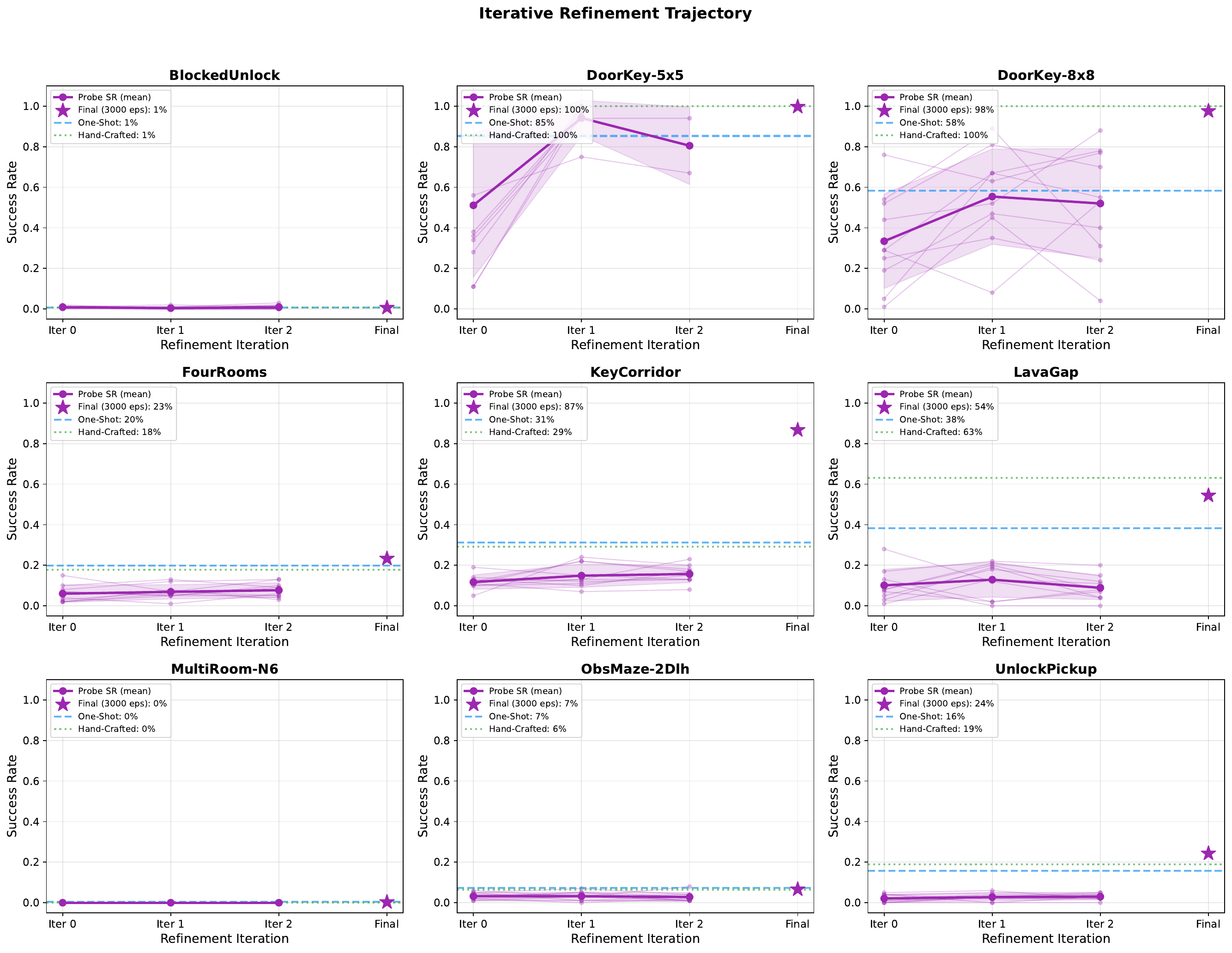}
  \caption{Refinement trajectory: probe success rate at each refinement round. Individual seed traces shown in light purple; mean in bold. Stars indicate final full-training results. Dashed lines show one-shot and hand-crafted baselines.}
  \label{fig:ablation}
\end{figure}

Figure~\ref{fig:ablation} shows how probe success rate evolves across refinement iterations. Key observations:

\begin{itemize}
  \item \textbf{DoorKey-5$\times$5}: Often solved within 1--2 iterations; all methods converge to $\sim$100\% given 3{,}000 episodes, making this primarily a sample-efficiency benchmark.
  \item \textbf{DoorKey-8$\times$8}: Probe performance improves across refinement rounds, and full training with the refined reward reaches $97.6\%$. LLM One-Shot reaches $58.3\%$, while hand-crafted shaping reaches $100\%$.
  \item \textbf{KeyCorridor}: LLM Iterative reaches $86.7\%$, compared with $31.2\%$ for LLM One-Shot and $29.2\%$ for hand-crafted shaping, but with high seed-to-seed variance. The static-vocabulary, prompt-matched, and \texttt{event\_text} controls show that this end-to-end gap cannot be attributed to dynamic diagnostic labels alone.
  \item \textbf{LavaGap}: Under corrected task-success accounting, LLM Iterative reaches $54.4\%$, above LLM One-Shot ($38.2\%$) but below no shaping ($64.4\%$) and hand-crafted shaping ($63.1\%$).
\end{itemize}

\subsection{Continuous Control Results}

\begin{figure}[H]
  \centering
  \includegraphics[width=\columnwidth]{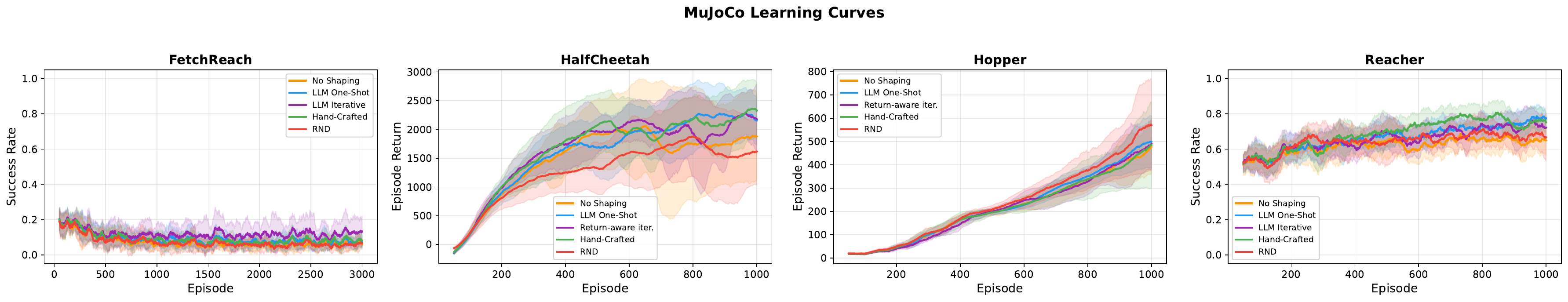}
  \caption{MuJoCo learning curves (smoothed over 50 episodes). Top: reaching tasks (success rate). Bottom: locomotion tasks (episode return). Shaded regions show $\pm$1 sample std across seeds; return-aware locomotion iterative curves use 6 seeds and other conditions use 10 seeds.}
  \label{fig:mujoco_results}
\end{figure}

To probe how the same reward-design interface behaves beyond discrete grid worlds, we evaluate on four MuJoCo environments (Figure~\ref{fig:mujoco_results}). The reward function interface remains identical---the LLM receives a natural-language environment description and generates a Python function---but the results are best read as boundary evidence rather than a broad claim about continuous control.

\textbf{Reaching tasks.} On Reacher-v4, LLM one-shot achieves the highest success rate ($77.0 \pm 4.1\%$), outperforming no shaping ($65.0\%$, Welch's $p < 0.001$, $d = 1.97$; significant with per-family correction within MuJoCo reaching). On FetchReach-v4 (3{,}000 episodes), iterative refinement achieves the highest success rate ($12.8 \pm 8.0\%$), roughly $2\times$ no shaping ($6.5\%$), though all conditions remain far from solving this challenging sparse-reward task.

\textbf{Locomotion tasks.} On HalfCheetah-v4 and Hopper-v4, the LLM designs alternative reward functions accessing velocity, position, and control cost. Unlike reaching tasks, these environments already have dense default rewards, making them a stress test for diagnostics designed around sparse binary success. In Table~\ref{tab:main_results}, the locomotion iterative column reports the return-aware diagnostic adaptation from Section~\ref{sec:locomotion_failure}, while the original success-diagnostic protocol is retained in Table~\ref{tab:locomotion_adaptation} as the broken iterative baseline. The main finding is negative: iterative refinement does not robustly outperform one-shot on locomotion tasks. On HalfCheetah, return-aware iterative refinement ($2{,}215 \pm 485$) is statistically indistinguishable from LLM one-shot ($2{,}190 \pm 405$, corrected $p=0.9795$). On Hopper, no condition significantly outperforms the baseline, and the budget-matched no-shaping control in Section~\ref{sec:budget_control} is much stronger than return-aware refinement. These results suggest that the feedback-driven diagnostic approach is most effective for sparse-reward environments where the LLM can identify specific failure modes.

\begin{figure}[H]
  \centering
  \includegraphics[width=\columnwidth]{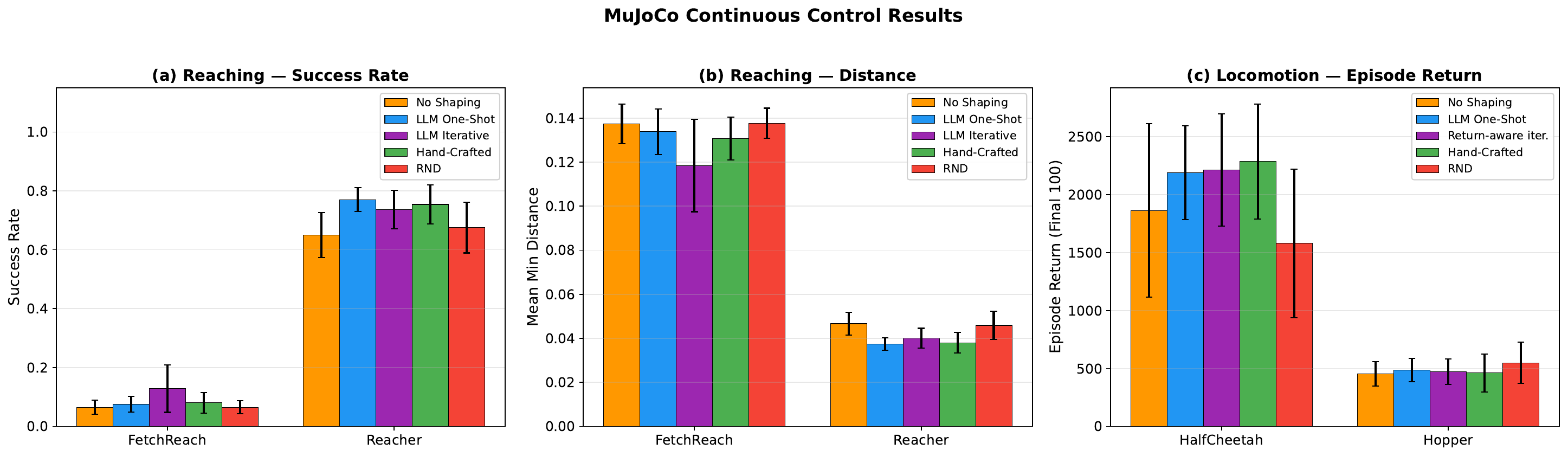}
  \caption{MuJoCo final performance: success rate for reaching tasks, episode return for locomotion tasks. Error bars show $\pm$1 sample std across seeds; return-aware locomotion iterative bars use 6 seeds and other conditions use 10 seeds.}
  \label{fig:mujoco_bars}
\end{figure}

\subsection{Secondary Comparisons}

We keep the main analysis focused on the diagnostic mechanism. Additional comparisons with RND, a naive LLM+RND combination, qualitative reward-function patterns, and protocol-level cost are summarized in Appendix~\ref{app:secondary}. These analyses provide useful context, but they are not the central evidence for the debugging thesis.

\subsection{Prompt-Matched Evaluation Across Backbones}
\label{sec:backbone_matched}

Table~\ref{tab:backbone_matched} evaluates whether refinement remains useful
when the one-shot generator is stronger. Within each row, informed one-shot and refinement share the exact initial reward function, static prompt, and RL seed.

\begin{table}[H]
\centering
\scriptsize
\setlength{\tabcolsep}{3pt}
\begin{tabular}{@{}llccccc@{}}
\toprule
\textbf{Backbone} & \textbf{Task} & \textbf{Informed 1-shot} &
\textbf{Refined} & \textbf{$\Delta$} & \textbf{95\% CI} &
\textbf{$p_{\rm Holm}$} \\
\midrule
Haiku 4.5 & DoorKey-8$\times$8 & $79.2 \pm 41.8$ & $94.6 \pm 6.3$ & $+15.4$ & $[-5.5,43.8]$ & $1.00$ \\
Haiku 4.5 & KeyCorridor & $52.1 \pm 44.3$ & $46.5 \pm 37.5$ & $-5.6$ & $[-40.3,30.5]$ & $1.00$ \\
Opus 5 & DoorKey-8$\times$8 & $91.1 \pm 28.1$ & $99.9 \pm 0.3$ & $+8.8$ & $[-0.3,26.7]$ & $1.00$ \\
Opus 5 & KeyCorridor & $84.2 \pm 21.8$ & $99.9 \pm 0.3$ & $+15.7$ & $[3.7,29.4]$ & $0.75$ \\
DeepSeek V4 Pro & DoorKey-8$\times$8 & $99.5 \pm 1.6$ & $99.6 \pm 0.7$ & $+0.1$ & $[-0.8,1.3]$ & $1.00$ \\
DeepSeek V4 Pro & KeyCorridor & $62.7 \pm 39.9$ & $77.9 \pm 37.4$ & $+15.2$ & $[-25.1,51.9]$ & $1.00$ \\
\bottomrule
\end{tabular}
\caption{Prompt-matched backbone control (final-100 success rate in percent, mean $\pm$ sample std; 10 paired seeds per row). $\Delta$ is refined minus informed one-shot. Confidence intervals use 10{,}000 paired bootstrap resamples; exact paired sign-flip tests are Holm-corrected across the six rows. No contrast survives the family correction.}
\label{tab:backbone_matched}
\end{table}

The effect is task- and model-dependent. DeepSeek V4 Pro is already at ceiling on DoorKey. Opus~5 has a $+15.7$-point KeyCorridor mean difference and a bootstrap interval above zero, but only four nonzero improvements and six ties, and its exact paired test does not survive correction. The remaining intervals include zero. Thus the experiment supports selective correction in some model--task pairs, not a uniform refinement gain.

All evaluated initial and final functions pass syntax and API lint, with no undocumented \texttt{info} fields or invalid action literals. DeepSeek nevertheless required 29 rejected intermediate refinement attempts before valid code was obtained; Appendix~\ref{app:secondary} reports the validation and cost details.

\subsection{Dense-Reward Boundary Case}
\label{sec:locomotion_failure}

The dense-reward locomotion result is primarily a boundary finding. The sharpest counterexample is Hopper: with the same total episode budget ($1{,}600$), no-shaping extended reaches $882$ while the original success-diagnostic iterative protocol reaches only $525$, and the return-aware adaptation remains lower at $472$ with a shorter 1{,}000-episode final training run. Thus the refinement loop provides no advantage when the primary bottleneck is training duration rather than reward-function quality. HalfCheetah is less negative but still not a robust gain: success-diagnostic iterative refinement achieves $2{,}040$ return compared to $2{,}190$ for one-shot, and return-aware refinement is statistically indistinguishable from one-shot.

\textbf{Root cause.} Investigation reveals a systematic false-positive failure in the diagnostic system. The reward-hacking detector fires when mean reward is high but success rate is low ($\text{mr} > 0.5 \wedge \text{sr} < 0.2$). On locomotion environments, \texttt{is\_success} is always \texttt{False} (there is no binary success criterion), so $\text{sr} = 0$ while $\text{mr} \sim 2{,}000$. This triggers a ``REWARD HACKING DETECTED'' warning \emph{every iteration}, causing the LLM to systematically strip effective shaping components from its reward function.

\textbf{Mechanism-removal diagnostic adaptation.} We disable success-based diagnostics for dense-reward environments and replace them with \emph{return-trend analysis}: the system compares mean return in the first vs.\ second half of the probe run, flagging ``RETURN DECLINING'' (second half $< 0.9 \times$ first half, indicating destabilizing shaping) or ``RETURN STAGNATED'' ($|\Delta| < 0.05 \times$ first half, indicating insufficient signal). A dedicated locomotion refinement prompt provides domain-specific guidance: velocity milestone bonuses, stability bonuses, and magnitude constraints ($<$10\% of typical episode return). This prompt omits success rate (always zero, confusing to the LLM) and instead frames diagnostics in terms of return trajectory.

We evaluate the adaptation on the first 6 completed canonical seeds per environment (Table~\ref{tab:locomotion_adaptation}). On HalfCheetah, the return-aware iterative method achieves $2{,}215 \pm 485$ mean return, numerically above the broken iterative baseline ($2{,}040 \pm 685$) and no shaping ($1{,}864 \pm 749$) but not distinguishable from one-shot. On Hopper, the adapted method ($472 \pm 112$) is comparable to no shaping ($453 \pm 104$) but below both the broken iterative baseline ($525 \pm 161$) and the budget-matched no-shaping control, consistent with Hopper benefiting primarily from training duration rather than reward function quality (Section~\ref{sec:budget_control}). Critically, the return-trend diagnostics \textbf{eliminate one specific systematic false-positive failure}: no run triggers the spurious ``REWARD HACKING'' warning that previously caused the LLM to strip effective shaping components.

\begin{table}[H]
\centering
\footnotesize
\setlength{\tabcolsep}{8pt}
\begin{tabular}{@{}lcccc@{}}
\toprule
\textbf{Env} & \textbf{No Shape} & \textbf{One-Shot} & \textbf{Broken Iter.} & \textbf{Return-Aware} \\
\midrule
HalfCheetah & $1864 \pm 749$ & $2190 \pm 405$ & $2040 \pm 685$ & $2215 \pm 485$ \\
Hopper      & $453 \pm 104$   & $485 \pm 101$   & $525 \pm 161$ & $472 \pm 112$ \\
\bottomrule
\end{tabular}
\caption{Dense-reward locomotion adaptation (mean $\pm$ sample std). Return-aware diagnostics remove the systematic false-positive reward-hacking warning, but do not establish robust locomotion gains. Return-aware results use the first 6 completed canonical seeds; other columns use 10 seeds.}
\label{tab:locomotion_adaptation}
\end{table}

This boundary case illustrates a central lesson: \textbf{iterative refinement diagnostics must be aligned with the task's evaluation signal}. In the sparse structured tasks studied here, where the diagnostics align with binary success signals, the framework is most effective. When they are misaligned (dense reward, no success criterion), the feedback loop becomes counterproductive. Return-trend analysis removes the specific false-positive reward-hacking mechanism in locomotion, but the Hopper result shows that removing one diagnostic failure mechanism is not enough to produce metric gains in dense continuous control.

\subsection{Episode Budget Control}
\label{sec:budget_control}

\begin{figure}[ht]
  \centering
  \includegraphics[width=\columnwidth]{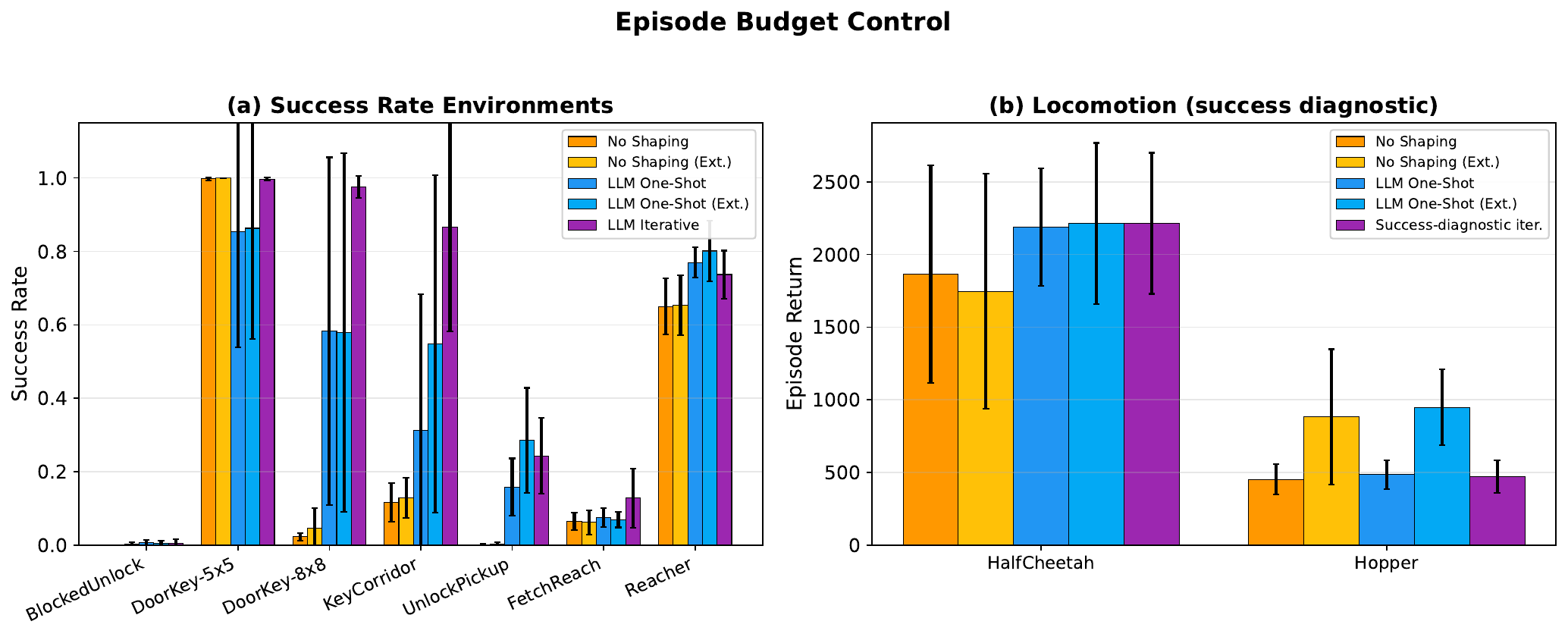}
\caption{Episode budget control: extended-budget conditions (4{,}500 episodes for MiniGrid) compared with standard and iterative conditions. MiniGrid extended-budget runs use 4{,}500 episodes; MuJoCo extended-budget runs use full budget $+$ 600 probe-equivalent episodes. Seed counts match the corresponding main/budget-control runs unless otherwise noted. For locomotion, the iterative bar uses the original success-diagnostic protocol, not the main Table~\ref{tab:main_results} return-aware adaptation; it is retained here as the broken diagnostic baseline.}
  \label{fig:extended_budget}
\end{figure}

The iterative protocol uses more total training episodes than one-shot (3$\times$500 probe + 3{,}000 full = 4{,}500 for MiniGrid; 3$\times$200 + full for MuJoCo). To control for this confound, we run extended-budget experiments where no-shaping and LLM one-shot conditions train for the same total episodes as iterative (Figure~\ref{fig:extended_budget}). On the hardest MiniGrid environments, iterative refinement's advantage persists strongly: DoorKey-8$\times$8 iterative ($97.6\%$) far exceeds both no-shaping extended ($4.7\%$; raw Welch $p < 10^{-16}$, $d = 21.3$) and LLM extended ($57.9\%$). KeyCorridor iterative ($86.7\%$) similarly outperforms no-shaping extended ($12.9\%$; raw Welch $p = 1.4 \times 10^{-5}$, $d = 3.6$) and LLM extended ($54.8\%$). These two extended no-shaping comparisons are part of the designated family in Appendix~\ref{app:primary_stats}. UnlockPickup iterative ($24.3\%$) also exceeds no-shaping extended ($0.3\%$; raw Welch $p < 10^{-4}$, $d = 3.3$), but we treat this budget-control comparison as descriptive rather than part of Table~\ref{tab:primary_stats}. These results show that additional training time alone does not explain the sparse-task performance of the end-to-end pipeline; the prompt-matched controls are needed to separate static guidance from corrective refinement. On Hopper, the budget-matched no-shaping control ($882$) far exceeds both standard no-shaping ($453$) and the original success-diagnostic iterative protocol ($525$). Even the return-aware adaptation remains below this matched-budget control ($472$ at 1{,}000 episodes vs.\ $882$ at 1{,}600 episodes), indicating that Hopper's bottleneck is training duration rather than reward design.

\subsection{Variance Decomposition}
\label{sec:variance_decomp}

To understand when iterative refinement is most useful, we decompose observed performance variability into two components: variability from LLM reward-function generation and variability from RL training stochasticity. We decompose these sources by fixing one and varying the other: (1)~fixing the LLM-generated reward function from seed~42 and training with all 10 RL seeds to approximate RL variance, and (2)~fixing RL seed~42 and training with each of the 10 LLM-generated reward functions to approximate LLM-generation variance. This should be interpreted as a \emph{point-estimate decomposition} rather than a formal variance partition: because each component is anchored to a single fixed seed, the two terms are not orthogonal and need not sum to $100\%$.

\begin{figure}[ht]
  \centering
  \includegraphics[width=\columnwidth]{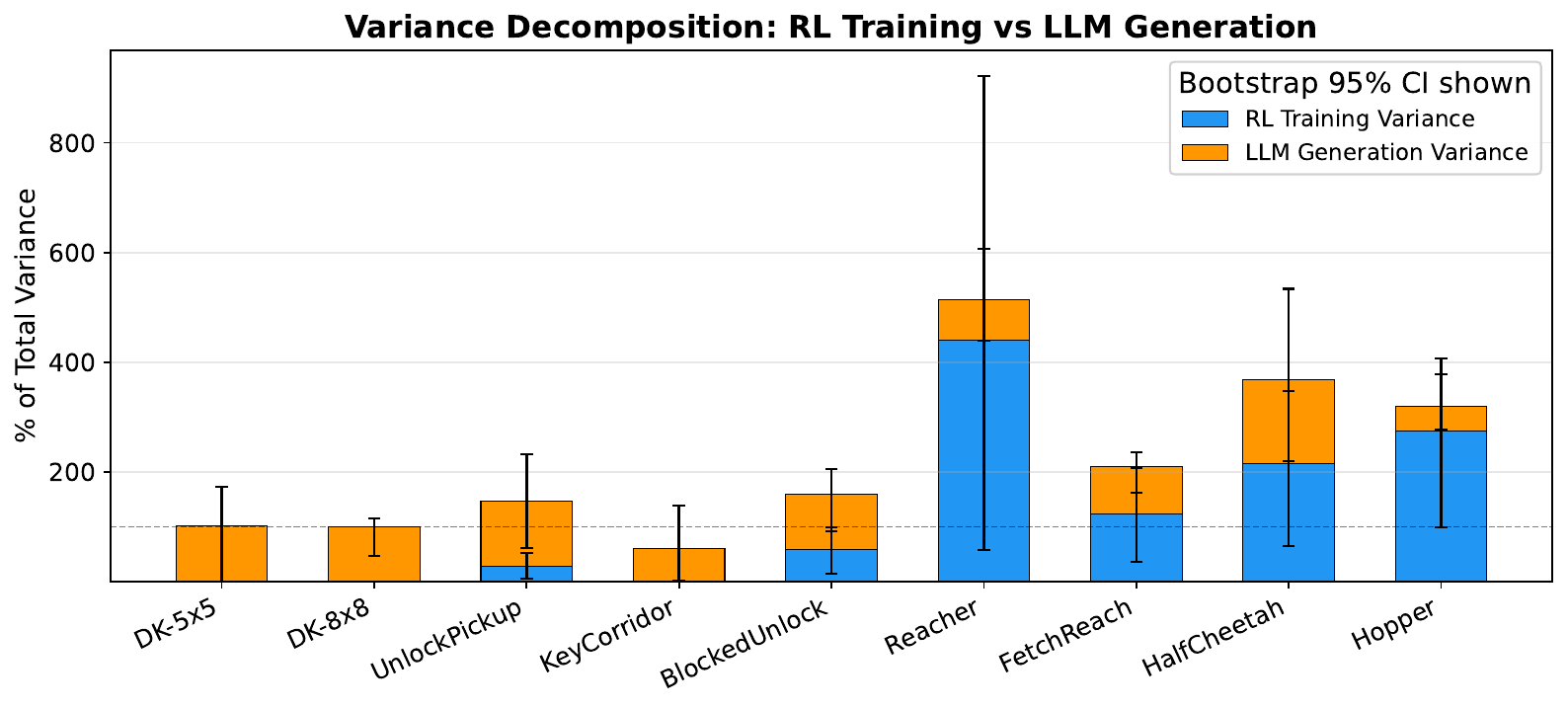}
  \caption{Anchored variance ratios for RL training vs.\ LLM generation per environment (single-seed anchored design). Ratios can exceed 100\% and are not variance shares because the components are non-orthogonal; see Table~\ref{tab:crossed_variance} for the fully crossed four-environment design.}
  \label{fig:variance_decomp}
\end{figure}

In the sparse, structured MiniGrid environments, LLM-function variance is typically much larger than RL variance. LLM variance point estimates range from $59$--$118\%$ of total variance, while RL variance is generally near zero (Figure~\ref{fig:variance_decomp}). DoorKey-8$\times$8 shows RL std of just $0.3\%$ vs.\ LLM std of $47.5\%$; DoorKey-5$\times$5 shows zero RL variance vs.\ $31.7\%$ LLM std. This suggests that performance is more sensitive to \emph{which} reward function the LLM produces than to RL training stochasticity, providing empirical support for why diagnostic refinement---improving a poor reward function---can materially change outcomes in these tasks.

In the continuous-control environments, the pattern shifts. RL variance point estimates are larger ($124$--$440\%$ of total variance), whereas LLM variance is more moderate ($44$--$153\%$). Although these ratios should not be interpreted literally as variance shares, they suggest that RL training stochasticity plays a larger role in continuous-control settings, consistent with the weaker gains from iterative refinement observed in the locomotion domain.

\textbf{Limitations of this decomposition.} Because each marginal variance is estimated using a single anchor seed rather than a fully crossed design, the components are not statistically independent, and their sum can exceed $100\%$ as observed in several environments. We report bootstrap $95\%$ confidence intervals ($2{,}000$ resamples, reproducible seed) alongside point estimates; these intervals are wide in several settings, especially continuous-control environments (e.g., HalfCheetah LLM std point estimate $499$, $95\%$ CI $[77, 722]$), indicating substantial instability in the exact ratio estimates. Accordingly, the anchored decomposition is best read as directional evidence about which source of variability appears more important in each regime, rather than as a precise estimate of variance explained.

\textbf{Fully crossed decomposition.} To complement the anchored analysis, we run a fully crossed $5 \times 5$ design (5~LLM reward functions $\times$ 5~RL seeds, all 25~combinations trained) on four representative environments spanning the variance spectrum. This supports a two-way random-effects variance decomposition with the standard estimator $\hat{\sigma}^2_{\text{LLM}} = (\text{MS}_{\text{LLM}} - \text{MS}_{\text{err}}) / R$ (and symmetrically for RL), clamping negative components to zero. Because there is one observation per LLM--RL cell, the residual term includes interaction effects and unmodeled stochasticity. Table~\ref{tab:crossed_variance} reports point-estimate variance shares and row/column-bootstrap intervals over the LLM and RL seed axes.

\begin{table}[H]
\centering
\footnotesize
\setlength{\tabcolsep}{8pt}
\begin{tabular}{@{}lcccc@{}}
\toprule
\textbf{Environment} & \textbf{$n$} & \textbf{LLM share} & \textbf{RL share} & \textbf{Residual} \\
\midrule
DoorKey-8$\times$8 & $25$ & $\mathbf{96.0\%}$ [$66.4,100.0$] & $0.0\%$ [$0.0,13.4$] & $4.0\%$ [$0.0,25.6$] \\
KeyCorridor        & $25$ & $\mathbf{50.2\%}$ [$0.0,89.8$] & $0.0\%$ [$0.0,51.1$] & $49.8\%$ [$6.3,94.4$] \\
Reacher-v4         & $25$ & $19.3\%$ [$0.0,69.8$] & $23.4\%$ [$0.0,100.0$] & $\mathbf{57.3\%}$ [$0.0,85.5$] \\
Hopper-v4          & $25$ & $7.3\%$ [$0.0,62.9$] & $30.1\%$ [$0.0,80.7$] & $\mathbf{62.6\%}$ [$12.8,93.2$] \\
\bottomrule
\end{tabular}
\caption{Crossed $5 \times 5$ variance decomposition via two-way random-effects ANOVA (25~cells per env, one observation per cell). Brackets show row/column-bootstrap $95\%$ intervals for variance shares, using the same negative-clamped estimator as the point estimates. The residual includes LLM--RL interaction effects and unmodeled stochasticity. \textbf{Bold} marks the dominant variance component. Metric: success rate for MiniGrid, $\text{final\_100\_reward}$ for MuJoCo.}
\label{tab:crossed_variance}
\end{table}

Table~\ref{tab:crossed_variance} suggests a domain split in the crossed design that the anchored decomposition could only approximate, but the bootstrap intervals are wide and should temper any fine-grained ranking. On the sparse discrete tasks, \textbf{LLM-function variance has the largest point estimate}: DoorKey-8$\times$8 attributes $96\%$ of variance to the LLM reward function identity (CI $[66,100]\%$), indicating that the identity of the LLM-generated reward function accounts for most observed performance variation in this setting. KeyCorridor shows a more balanced and uncertain split ($50\%$ LLM, CI $[0,90]\%$; $50\%$ residual, CI $[6,94]\%$), suggesting that LLM--RL interaction effects may play a larger role on this harder task. On the continuous-control tasks, the residual share has the largest point estimate ($57$--$63\%$), but the LLM and RL component intervals are broad. The anchored design's $275.6\%$ and $439.5\%$ RL shares on these envs (Figure~\ref{fig:variance_decomp}) were inflated by its non-orthogonal single-seed anchoring; the crossed design provides cleaner descriptive point estimates, not precise variance-share claims.

This decomposition supports the method's scoped interpretation most clearly for DoorKey-8$\times$8: iterative refinement---which improves the LLM reward function---should be most useful where LLM variance appears dominant. For KeyCorridor and the continuous-control tasks, the crossed estimates are better read as directional diagnostics than as settled component rankings.

\subsection{Selection vs.\ Refinement}
\label{sec:bestofn}

An alternative to iterative refinement is \emph{selection}: generate $N$ independent reward functions, evaluate each via a short probe, and train with the best. We implement best-of-3 selection with the same total episode budget as iterative (3$\times$500 probe + 3{,}000 full for MiniGrid). This approach captures some benefit of variance reduction by selecting the best from multiple independent samples.

\begin{figure}[ht]
  \centering
  \includegraphics[width=\columnwidth]{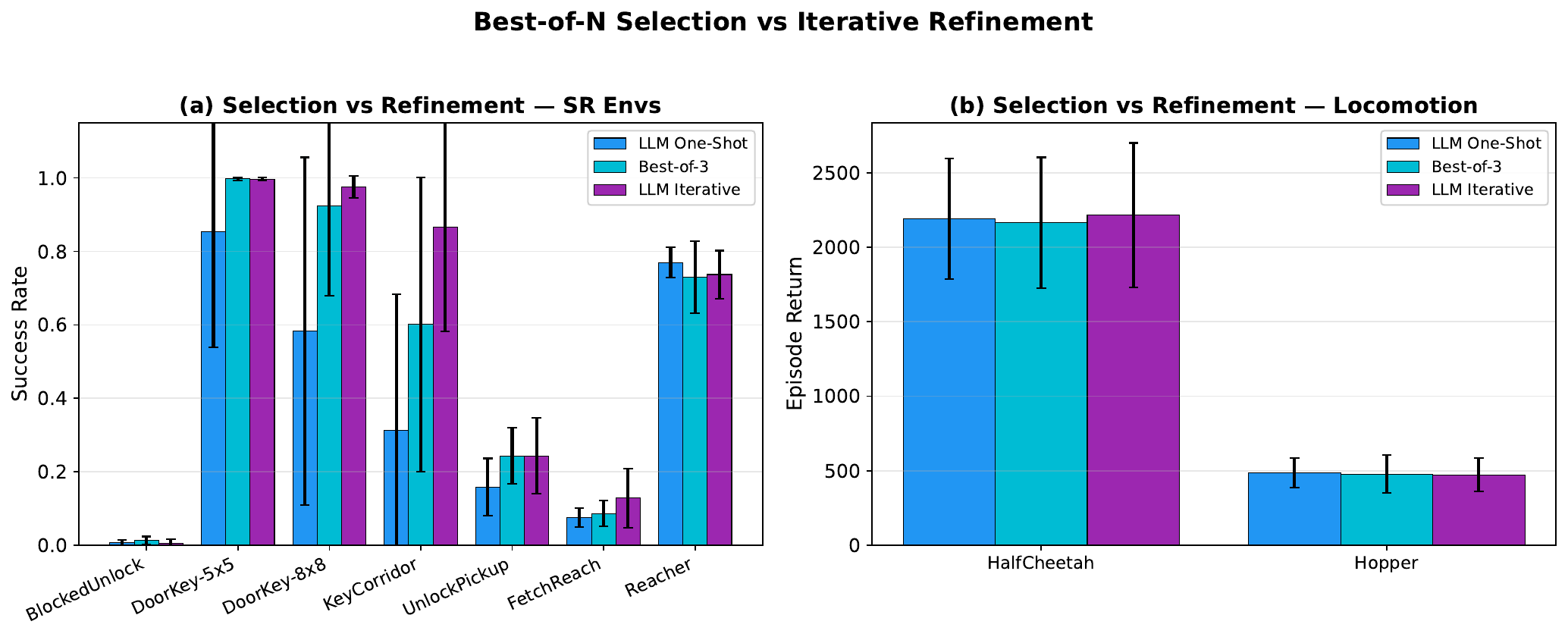}
  \caption{Best-of-3 selection vs.\ LLM one-shot vs.\ iterative refinement. MiniGrid Best-of-3 uses 10 seeds; MuJoCo locomotion Best-of-3 has incomplete validation-successful seeds (HalfCheetah: 5/10, Hopper: 4/10).}
  \label{fig:bestofn}
\end{figure}

Figure~\ref{fig:bestofn} compares best-of-3 selection with one-shot and iterative refinement. On DoorKey-8$\times$8, best-of-3 ($92.3\%$) substantially improves over one-shot ($58.3\%$) by filtering out poor candidates, though iterative ($97.6\%$) still achieves the highest mean performance. On KeyCorridor, best-of-3 ($60.1\%$) roughly doubles one-shot ($31.2\%$) but remains below iterative ($86.7\%$). On UnlockPickup, best-of-3 ($24.3\%$) matches iterative ($24.3\%$), both improving over one-shot ($15.8\%$). Seed-level tests do not resolve the iterative-vs-selection differences at $n=10$: DoorKey-8$\times$8 $\Delta=5.3$ points (Welch $p=0.51$, bootstrap CI $[-3.8,21.4]$), KeyCorridor $\Delta=26.6$ points ($p=0.106$, CI $[-2.6,54.8]$), and UnlockPickup $\Delta=0.0$ points ($p=1.0$, CI $[-7.9,6.9]$). We therefore interpret Best-of-3 as a strong selection baseline: it captures most of the DoorKey-8$\times$8 gain and matches UnlockPickup, while the KeyCorridor mean remains descriptively lower. Note that best-of-3 MuJoCo locomotion results are based on fewer seeds (HalfCheetah: 5/10, Hopper: 4/10) due to reward function validation failures, limiting their statistical reliability.

\paragraph{Call- and episode-budget-matched population refinement.}
We additionally implement a shared-harness population control inspired by population-based reward search. With Haiku and the informed prompt, it generates and probes two candidates, selects by raw task success and return, produces and probes two metrics-guided revisions without dynamic diagnostic labels, and full-trains the best function. Each candidate or revision receives a 375-episode probe, so the four LLM calls and 4{,}500 total episodes match the maximum diagnostic-refinement budget.

\begin{table}[H]
\centering
\small
\setlength{\tabcolsep}{5pt}
\begin{tabular}{@{}lccccc@{}}
\toprule
\textbf{Task} & \textbf{Diagnostic} & \textbf{Population} &
\textbf{$\Delta$} & \textbf{95\% CI} & \textbf{$p$} \\
\midrule
DoorKey-8$\times$8 & $94.6 \pm 6.3$ & $98.4 \pm 2.1$ &
$-3.8$ & $[-8.1,0.1]$ & $.150$ \\
KeyCorridor & $46.5 \pm 37.5$ & $61.7 \pm 41.1$ &
$-15.2$ & $[-43.0,12.7]$ & $.336$ \\
\bottomrule
\end{tabular}
\caption{Haiku diagnostic refinement versus the four-call population-refinement control (final-100 success rate in percent, mean $\pm$ sample std; 10 paired seeds). $\Delta$ is diagnostic minus population. Intervals use 10{,}000 paired bootstrap resamples; $p$ is the exact paired sign-flip value.}
\label{tab:population_control}
\end{table}

The population control has the higher mean on both tasks, although neither paired difference is resolved. The two protocols are therefore competitive under this shared budget; this is a within-harness control, not an official Eureka reproduction or a cross-benchmark comparison.

\section{Related Work}
\label{sec:related}

\textbf{LLMs as reward designers.}
Eureka~\citep{ma2023eureka} uses GPT-4 to generate reward functions for continuous-control tasks, employing evolutionary search over hundreds of candidates evaluated in GPU-parallel simulation (IsaacGym). Text2Reward~\citep{xie2023text2reward} generates dense executable reward code from language task descriptions and compact environment representations. ELLM~\citep{du2023guiding} uses LLMs to suggest exploration goals rather than reward functions. Language to Rewards~\citep{yu2023language2reward} translates task specifications into reward parameters for manipulation. More recent work extends this paradigm: Auto MC-Reward~\citep{li2024automcreward} applies automated dense reward design to Minecraft, CARD~\citep{sun2025card} refines LLM-generated reward code using dynamic feedback and trajectory preference evaluation, and RL-VLM-F~\citep{wang2024rl} uses vision-language models for reward feedback. \citet{kwon2023reward} study prompting LLMs as proxy reward functions from zero- or few-shot task descriptions. Our shared-harness population control compares search and diagnosis using the same LLM, interface, PPO trainer, call count, and episode budget (Table~\ref{tab:population_control}); because the official systems use different simulators and protocols, we do not present it as an official Eureka or Text2Reward reproduction.

\textbf{Reward misspecification and hacking.}
A growing literature studies the risks of misspecified rewards. \citet{amodei2016concrete} identify reward hacking as a key AI safety concern. \citet{skalse2022defining} provide a formal characterization of reward hacking. \citet{pan2022effects} map the effects of reward misspecification on learned policies. \citet{clark2016faulty} document faulty reward functions in deployed systems. Our failure-mode taxonomy (reward flooding, semantic/API mismatch under basic prompts, and rarer weak-shaping cases) connects to this literature by characterizing the specific ways LLM-generated rewards become misspecified, and our diagnostic protocol provides a lightweight way to flag these failure modes during refinement.

\textbf{Validity of diagnostic signals.}
Chain-of-thought explanations can omit factors that influence model outputs~\citep{turpin2023language}, while interventions on reasoning traces can test how answer generation depends on those traces~\citep{lanham2023measuring}. In tutoring with a supplied answer key, \citet{Shen2026ContextMasked} find that early-answer signals in truncated reasoning audits can largely reflect answer-key information retained in the probe context rather than information in the explanation prefix. At the benchmark level, \citet{Wang2026SafetyCapability} show that the interpretation of agent-safety scores can depend on the metric, target behavior, and evaluated model panel. Although these settings differ from reward design, they highlight the need for controls when interpreting diagnostic signals. Our prompt controls and threshold sensitivity analyses examine the utility and calibration limits of reward-function diagnostics, while dense-reward boundary tests expose false positives when success-based rules are applied to locomotion tasks without a binary success criterion.

\textbf{Reward shaping theory.}
Potential-based reward shaping~\citep{ng1999policy} provides theoretical guarantees on policy invariance under additive shaping. \citet{devlin2012dynamic} extend this to dynamic potentials. \citet{wiewiora2003principled} establish connections between potential-based shaping and Q-value initialization. \citet{hu2020learning} propose learning to utilize shaping rewards adaptively. Our generated rewards are heuristic and do not generally satisfy the potential-based form; Section~\ref{sec:interpretation} uses this theory only to clarify that distinction.

\textbf{Intrinsic motivation.}
Count-based exploration~\citep{bellemare2016unifying}, curiosity-driven methods~\citep{pathak2017curiosity}, and Random Network Distillation~\citep{burda2019exploration} provide intrinsic rewards without domain knowledge. \citet{oudeyer2007intrinsic} survey intrinsic motivation systems for autonomous development. Go-Explore~\citep{ecoffet2021first} uses archiving and goal-directed exploration. \citet{taiga2021bonus} provide a systematic evaluation of bonus-based exploration. We include fixed-coefficient RND as a contextual baseline and add a small DoorKey coefficient sweep. The sweep confirms that RND is highly coefficient-sensitive: tuning solves DoorKey-5$\times$5 but does not rescue DoorKey-8$\times$8 under our PPO protocol.

\textbf{LLMs for code generation.}
Code-generating LLMs~\citep{chen2021evaluating} have been applied broadly, with specialized models like StarCoder~\citep{li2023starcoder} and Code Llama~\citep{roziere2023code} achieving strong performance on programming benchmarks. In RL, this capability enables generating executable reward functions~\citep{ma2023eureka}, embodied-agent skill code~\citep{wang2023voyager}, and policy code~\citep{liang2023code}. Our iterative refinement with training diagnostics is analogous to the code-test-debug cycle in software engineering, but applied to reward function design rather than general programming.

\section{Limitations and Future Work}
\label{sec:limitations}

\textbf{Scope: structured interfaces required.} Our method depends on a curated reward-function interface: the \texttt{info} dictionary exposes semantic fields such as \texttt{carrying} and \texttt{agent\_pos}. An interface ablation (Section~\ref{sec:scope_ablation}) shows that the method does not critically depend on \texttt{event\_text}, but it does require some structured semantic interface. The present experiments therefore do not establish applicability to RLHF, open-ended LLM agents, raw-pixel embodied tasks, or human-preference rewards, where the relevant signals and failure modes may be unknown. We also cannot determine whether the pretrained backbones encountered MiniGrid descriptions or code; the matched experiment tests use of a supplied interface, not generalization to genuinely novel task families.

\textbf{Dense-reward settings require different diagnostics.} As demonstrated in Section~\ref{sec:locomotion_failure}, success-based heuristics are counterproductive on dense-reward tasks. Our return-trend adaptation addresses the specific locomotion failure mechanism without producing a robust performance gain. Other dense-reward domains may require further diagnostic adaptations.

\textbf{Diagnostic thresholds.} The diagnostics use fixed heuristic thresholds (Table~\ref{tab:thresholds}), not universal definitions of reward failure. The sensitivity analysis (Section~\ref{sec:threshold_sensitivity}) shows task-dependent effects, including for the $0.2$ success-rate bound, and the component-removal stress test exposes false positives. Adaptive calibration based on environment characteristics or early training dynamics is a natural direction for future work.

\textbf{Environment complexity.} While we evaluate across 13 environments spanning discrete grid worlds, continuous reaching, and continuous locomotion, these use low-dimensional state observations. Extending to high-dimensional observation spaces and long-horizon manipulation tasks would test the scope of the approach beyond the structured setting studied here.

\textbf{Scaling to harder tasks.} BlockedUnlockPickup remains below $1\%$ success across all methods, MultiRoom-N6 near $0\%$, and UnlockPickup achieves only $\sim$24\% with iterative refinement. Tasks requiring very long planning horizons or complex reward decompositions remain challenging.

\textbf{Model diversity.} The prompt-matched study evaluates Haiku~4.5, Opus~5, and DeepSeek V4 Pro on two tasks with 10 paired seeds. This addresses dependence on Haiku but does not establish a general model ranking: effects remain task- and model-dependent, and none of the six contrasts survives family-wise correction. Our DeepSeek runs use the hosted endpoint; prompts, returned model text, generated code, and validator outcomes are archived for reproducibility.

\textbf{Algorithm and baseline scope.} All experiments use PPO, so some observed failure modes may reflect interactions between reward design and PPO optimization rather than reward functions alone. The main RND baseline uses a fixed intrinsic-reward coefficient of $0.1$; a DoorKey sweep shows that tuning changes the conclusion on DoorKey-5$\times$5 but not DoorKey-8$\times$8. The population-refinement control is a shared-harness adaptation, not an official head-to-head evaluation of Eureka or Text2Reward.

\textbf{Variance decomposition.} The anchored decomposition (Section~\ref{sec:variance_decomp}) uses a single seed per component, producing non-orthogonal point estimates. We supplement it with a fully crossed $5 \times 5$ design on four representative environments (Table~\ref{tab:crossed_variance}), but its bootstrap intervals remain wide. Extending the crossed design beyond these four environments would strengthen the analysis.

\textbf{Refinement protocol design.} The current protocol discards the probe-trained policy and retrains from scratch, which simplifies evaluation but wastes probe computation. Strong warning labels may also prime the LLM toward overcorrection. Warm-starting and softer or more specific feedback are important future directions.

\textbf{Model settings and per-seed LLM calls.} Haiku uses temperature $0.4$. Opus~5 uses the high-effort setting, and DeepSeek V4 Pro uses thinking mode; the exact provider configurations are included in the supplement. Each seed uses its own LLM-generated reward function, conflating LLM and RL variance; our decomposition analysis quantifies but does not eliminate this confound.

\section{Conclusion}
\label{sec:conclusion}

We have argued that LLM reward design for sparse, structured RL tasks is better understood as a debugging problem than a generation problem. One-shot generation fails in recurring, identifiable ways; lightweight diagnostics can sometimes detect and repair these failures in the studied sparse structured setting; and the approach breaks predictably when diagnostics are misaligned with the task's evaluation signal.

The end-to-end pipeline achieves $97.6\%$ success on DoorKey-8$\times$8 and $86.7\%$ on KeyCorridor, though KeyCorridor remains high variance. Component-removal tests, metrics-only re-prompting, and the static-vocabulary control show that the taxonomy is useful but that dynamic warning labels are not the sole mechanism. The stricter prompt-matched study further narrows the conclusion: effects vary across tasks and backbones, none of the six contrasts survives family-wise correction, and a four-call population-refinement control is competitive. We therefore interpret the contribution as taxonomy-guided prompting with diagnostic correction, not uniform superiority of iterative refinement over informed one-shot generation or population search.

A fully crossed $5 \times 5$ variance decomposition provides descriptive support for when refinement may be useful: on DoorKey-8$\times$8, LLM-function variance has a $96\%$ point-estimate share with a wide bootstrap interval. On continuous-control tasks, residual and RL variance have larger point estimates, limiting the benefit of improving any single reward function. The locomotion boundary case similarly removes one false-positive mechanism without producing robust gains.

The method is strongest for sparse, structured tasks with reliable semantic interfaces, and all experiments use PPO-trained policies. Richer fields such as \texttt{event\_text} are not uniformly helpful, threshold effects are task-dependent, and the current experiments do not establish transfer to less structured reward-design settings.

\bibliographystyle{tmlr}
\bibliography{references}

\appendix

\section{Secondary Analyses and Practical Takeaways}
\label{app:secondary}

\textbf{Intrinsic motivation.} With the fixed coefficient used in the main table, RND behaves differently from LLM-designed shaping. Under corrected task-success accounting, it reaches $52.3\%$ on LavaGap and does not improve over no shaping ($64.4\%$). The DoorKey sweep also shows high coefficient sensitivity: $c=0.01$ solves DoorKey-5$\times$5, while $c=0.1$ and $c=0.5$ do not, and none of the three coefficients solves DoorKey-8$\times$8. A naive LLM+RND additive combination also hurts structured MiniGrid tasks (e.g., DoorKey-8$\times$8 drops from $58.3\%$ with LLM alone to $0.1\%$), suggesting that task-specific and domain-agnostic rewards require careful balancing rather than simple addition.

\textbf{Evolutionary reward design and cost.} Population-based methods such as Eureka evaluate many reward candidates in GPU-parallel simulation. Our four-call population-refinement control instead uses the same MiniGrid interface, Haiku backbone, PPO implementation, and 4{,}500-episode budget as diagnostic refinement. It is competitive and has higher means on both tested tasks (Table~\ref{tab:population_control}). This supports a within-harness comparison between two low-call protocols, not a performance claim against the official Eureka system.

\textbf{Qualitative reward-function patterns.} Generated rewards commonly use event-driven one-time bonuses, position-based progress heuristics, and bonus magnitudes comparable to hand-crafted shaping after refinement. These qualitative patterns support the debugging interpretation but are secondary to the controlled comparisons in the main text.

\textbf{Backbone validation and API usage.} The prompt-matched 10-seed evaluation in Table~\ref{tab:backbone_matched} yields 80 accepted Haiku artifacts and 76 accepted Opus~5 artifacts with no rejected attempts; two Opus runs stop early under the prespecified success rule. DeepSeek V4 Pro yields 80 accepted artifacts and 29 rejected intermediate attempts: 14 function-extraction failures, 11 disallowed imports, 3 invalid return contracts, and 1 undefined \texttt{np}. The archived API ledger records 466 attempts and 462 successful responses. The estimated total cost is USD~9.49 based on token usage and model prices; it is not an invoice total.

\textbf{Practical takeaways.} One-shot generation is most reliable on straightforward tasks with clear subgoals; corrective refinement may be useful when generated rewards remain poor or high variance; Best-of-$N$ selection is useful when independent candidates include a good reward; dense-reward tasks require diagnostic adaptation; and the variance analysis suggests prioritizing reward-function quality when LLM variance dominates, while prioritizing RL optimization when training stochasticity dominates.

\section{Primary Statistical Tests}
\label{app:primary_stats}

Table~\ref{tab:primary_stats} lists 12 designated comparisons from the broad evaluation. The comparison set focuses on sparse structured tasks, includes two sparse-task budget controls (DoorKey-8$\times$8 and KeyCorridor LLM Iterative vs.\ extended no shaping), and treats MuJoCo reaching and locomotion as separate families. The set was not externally preregistered and does not include every cell in Table~\ref{tab:main_results} or every descriptive statistic in Section~\ref{sec:budget_control}. We apply Welch's $t$-test within each environment and Holm--Bonferroni correction separately within the MiniGrid, MuJoCo reaching, and MuJoCo locomotion families. The table is generated from the same logs as Table~\ref{tab:main_results}; locomotion uses episode return and all other rows use final-100-episode success rate.

\begin{table}[H]
\centering
\scriptsize
\setlength{\tabcolsep}{3pt}
\begin{tabular}{@{}llccc@{}}
\toprule
\textbf{Family / Env} & \textbf{Comparison} & \textbf{$\Delta$} & \textbf{$p_{\mathrm{corr}}$} & \textbf{Sig.} \\
\midrule
MiniGrid / DoorKey-8$\times$8 & Iterative vs.\ No Shaping & $+95.3$ pp & $<0.0001$ & yes \\
MiniGrid / KeyCorridor & Iterative vs.\ No Shaping & $+75.0$ pp & $0.0001$ & yes \\
MiniGrid / UnlockPickup & Iterative vs.\ No Shaping & $+24.2$ pp & $0.0001$ & yes \\
MiniGrid / DoorKey-8$\times$8 & Iterative vs.\ One-Shot & $+39.3$ pp & $0.0275$ & yes \\
MiniGrid / KeyCorridor & Iterative vs.\ One-Shot & $+55.5$ pp & $0.0032$ & yes \\
MiniGrid / DoorKey-8$\times$8 & Iterative vs.\ Extended No Shaping & $+92.9$ pp & $<0.0001$ & yes \\
MiniGrid / KeyCorridor & Iterative vs.\ Extended No Shaping & $+73.8$ pp & $0.0001$ & yes \\
\midrule
MuJoCo Reaching / Reacher & One-Shot vs.\ No Shaping & $+12.0$ pp & $0.0012$ & yes \\
MuJoCo Reaching / FetchReach & Iterative vs.\ No Shaping & $+6.3$ pp & $0.0373$ & yes \\
\midrule
MuJoCo Locomotion / HalfCheetah & One-Shot vs.\ No Shaping & $+326$ return & $0.7395$ & no \\
MuJoCo Locomotion / Hopper & One-Shot vs.\ No Shaping & $+32$ return & $0.9795$ & no \\
MuJoCo Locomotion / HalfCheetah & Return-Aware Iterative vs.\ One-Shot & $+25$ return & $0.9795$ & no \\
\bottomrule
\end{tabular}
\caption{Designated comparisons from the broad evaluation. Significance is assessed at $p_{\mathrm{corr}} < 0.05$ after per-family Holm--Bonferroni correction. pp = percentage points.}
\label{tab:primary_stats}
\end{table}

\section{Threshold Definitions}
\label{app:thresholds}

Table~\ref{tab:thresholds} enumerates all trigger values used in the iterative refinement procedure. All values correspond exactly to those used in our implementation.

\begin{table}[H]
\centering
\footnotesize
\setlength{\tabcolsep}{6pt}
\begin{tabular}{@{}llll@{}}
\toprule
\textbf{Category} & \textbf{Component} & \textbf{Value} & \textbf{Applies To} \\
\midrule
\multicolumn{4}{l}{\textit{Refinement Control}} \\
& Success threshold $\tau$ & $0.95$ & Binary-success tasks \\
& Max iterations $K$ & $3$ & All environments \\
& Probe episodes $N_p$ & $500$ & MiniGrid \\
& Probe episodes $N_p$ & $200$ & MuJoCo \\
& Full episodes $N_f$ & $3{,}000$ & MiniGrid, FetchReach \\
& Full episodes $N_f$ & $1{,}000$ & Reacher, HalfCheetah, Hopper \\
\midrule
\multicolumn{4}{l}{\textit{Diagnostics (Sparse-Reward)}} \\
& High-return/low-success flag & $\text{mr} > 0.5\ \wedge\ \text{sr} < 0.2$ & MiniGrid, reaching \\
& Shaping weakness & $\text{mr} < 0.1\ \wedge\ \text{sr} < 0.1$ & MiniGrid, reaching \\
& Learning plateau & $0.1 < \text{sr} < 0.7\ \wedge\ \text{eps} > 1000\ \wedge\ \Delta\text{sr} < 0.05$ & Longer MiniGrid probes \\
\midrule
\multicolumn{4}{l}{\textit{Diagnostics (Dense-Reward)}} \\
& Reward hacking & Disabled & Locomotion \\
& Return declining & $\bar{r}_2 < 0.9 \times \bar{r}_1$ & Locomotion \\
& Return stagnated & $|\bar{r}_2 - \bar{r}_1| < 0.05 \times \bar{r}_1$ & Locomotion \\
\midrule
\multicolumn{4}{l}{\textit{Evaluation}} \\
& Success criterion & distance $< 0.05$ & Reacher, FetchReach \\
& Sample efficiency & $80\%$ success rate & Figure~\ref{fig:sample_efficiency} \\
\bottomrule
\end{tabular}
\caption{All trigger values and thresholds used in the iterative refinement procedure. No hidden defaults exist beyond this table.}
\label{tab:thresholds}
\end{table}

\section{Prompt Templates}
\label{app:prompts}

\subsection{Generation Prompt}

The following compact template is used by the LLM One-Shot condition in the broad evaluation. The \texttt{\{env\_description\}} placeholder is filled with a natural-language description of the environment. The prompt-matched control below uses a complete interface specification and identical static guidance across conditions.

\begin{lstlisting}[style=python, basicstyle=\ttfamily\tiny]
You are an expert reward function designer for RL.

Given a description of an environment, write a
Python reward shaping function that helps a PPO
agent learn faster.

# ENVIRONMENT
{env_description}

# FUNCTION SIGNATURE
def reward_fn(obs, action, reward, terminated,
              truncated, info, state):
    # obs: numpy array (7,7,3)
    # action: int (0=left, 1=right, 2=forward,
    #              3=pickup, 4=drop, 5=toggle)
    # info: dict with "event_text", "carrying",
    #        "agent_pos", "step_count", "max_steps"
    # state: mutable dict, reset each episode
    # Returns: (shaped_reward, state)

# DESIGN PRINCIPLES
1. Start with the original reward.
2. Add one-time bonuses (+0.1 to +0.3) for subgoals.
3. Use state dict to ensure bonuses given only ONCE.
4. Check info["event_text"] for events.
5. Keep bonuses small vs goal reward (~1.0).
6. No imports. Self-contained function.
\end{lstlisting}

\subsection{API-Complete Prompt for the Matched Controls}

The informed one-shot and paired refinement conditions use the same task description followed by the contract below. The exact rendered prompts are included in the supplementary artifacts.

\begin{lstlisting}[style=python, basicstyle=\ttfamily\tiny]
# EXECUTABLE INTERFACE
def reward_fn(obs, action, reward, terminated,
              truncated, info, state):
    return shaped_reward, state

obs has shape (7, 7, 3) and is the partial
egocentric view after the action:
  obs[x,y,0]: object ID
    0=unseen, 1=empty, 2=wall, 3=floor,
    4=door, 5=key, 6=ball, 7=box,
    8=goal, 9=lava, 10=agent
  obs[x,y,1]: color ID
    0=red, 1=green, 2=blue, 3=purple,
    4=yellow, 5=grey
  obs[x,y,2]: object-state ID
    0=open, 1=closed, 2=locked
The first two axes are local-view coordinates.

action is the integer just taken:
  0=turn left, 1=turn right, 2=move forward,
  3=pick up, 4=drop, 5=toggle, 6=done

Only these info fields are guaranteed:
  text_state: str
  action_text: str
  event_text: str
  carrying: str
  agent_pos: tuple[int, int]
  agent_dir: int (0=right,1=down,2=left,3=up)
  carrying_type: str or None
  step_count: int
  max_steps: int

state is persisted within an episode and reset to
{} on reset. Return a finite numeric shaped_reward
and a dict state. Do not import modules or use
undocumented fields.

# REWARD-DESIGN CHECKS
1. Avoid repeated reward accumulation. Prefer
   one-time milestones recorded in state.
2. Follow the documented interface and action
   semantics; actions 0 and 1 are rotations.
3. Ensure shaping conditions are reachable and
   non-negligible, but subordinate to task success.
\end{lstlisting}

The complete static prompt---task description, interface, reward-design checks, and implementation requirements---is byte-identical across the informed one-shot and paired refinement conditions. Refinement then appends the current source, probe results, and dynamically triggered feedback.

\subsection{Refinement Prompt}

In the broad evaluation, LLM Iterative appends the previous reward function, training metrics, diagnosed failure modes, and the following static reminder. The prompt-matched experiment instead uses the API-complete template above.

\begin{lstlisting}[style=python, basicstyle=\ttfamily\tiny]
# CURRENT REWARD FUNCTION
{current_source}

# TRAINING RESULTS
- Episodes trained: {episodes_trained}
- Success rate: {success_rate}
- Mean reward: {mean_reward}

# DIAGNOSED ISSUES
{diagnoses}

# COMMON FAILURE MODES TO AVOID
1. Reward flooding: Do NOT add per-step bonuses.
2. Action-index confusion: MiniGrid action
   0=turn_left, 2=forward, NOT directions.
3. Too-weak shaping: +0.1 may be too small.
   Use position-based progress tracking.
\end{lstlisting}

\section{Iterative Refinement Example}
\label{app:example}

Table~\ref{tab:refinement_example} shows the iterative refinement process on DoorKey-8$\times$8 (seed 42). The LLM progressively improves the reward function based on probe training diagnostics.

\begin{table}[ht]
\centering
\small
\begin{tabular}{@{}clp{5cm}@{}}
\toprule
\textbf{Iter} & \textbf{Probe SR} & \textbf{Key Changes} \\
\midrule
0 & 29\% & Basic one-time bonuses for key pickup (+0.2) and door open (+0.25). Continuous holding-key bonus causes reward flooding. \\
1 & 67\% & Removed continuous bonus. Added position-based progress: bonus for reaching new max $x$-coordinate. Tracks exploration state. \\
2 & 78\% & Raised milestone bonuses to +0.5. Added distance- and positive-$x$ exploration bonuses. \\
\midrule
\multicolumn{2}{l}{Final (3000 eps)} & \textbf{99\%} success rate \\
\bottomrule
\end{tabular}
\caption{Iterative refinement on DoorKey-8$\times$8 seed 42. Each iteration receives the previous function's code and probe metrics. The LLM diagnoses issues and revises the reward function, improving from 29\% to 99\%. See Appendix~\ref{app:code_examples} for the actual generated code.}
\label{tab:refinement_example}
\end{table}

The progression shows the LLM addressing specific weaknesses: iteration~1 adds exploration incentives when basic milestones prove insufficient, and iteration~2 increases the milestone and exploration bonuses. The final revision then removes the continuous exploration bonuses and uses room-aware progress only to reset a stagnation counter.

\section{LLM-Generated Reward Functions}
\label{app:code_examples}

We present the actual LLM-generated reward functions from the DoorKey-8$\times$8 iterative refinement (seed 42), showing how the model evolves its reward design across iterations.

\subsection{Iteration 0: Initial One-Shot Generation (Probe SR: 29\%)}

The initial function uses basic event-driven bonuses but includes two problematic elements: a continuous per-step reward for holding the key (\texttt{+0.01}) that can cause reward flooding, and a late-game penalty that discourages exploration.

\begin{lstlisting}[style=python, basicstyle=\ttfamily\tiny]
def reward_fn(obs, action, reward, terminated,
              truncated, info, state):
    shaped_reward = reward
    if "key_picked_up" not in state:
        state["key_picked_up"] = False
    if "door_opened" not in state:
        state["door_opened"] = False
    if "goal_reached" not in state:
        state["goal_reached"] = False
    event_text = info.get("event_text", "")
    carrying = info.get("carrying", "")
    step_count = info.get("step_count", 0)
    max_steps = info.get("max_steps", 1)
    # One-time bonuses
    if not state["key_picked_up"] \
       and "picked up" in event_text.lower() \
       and "key" in event_text.lower():
        shaped_reward += 0.2
        state["key_picked_up"] = True
    if not state["door_opened"] \
       and "opened door" in event_text.lower():
        shaped_reward += 0.25
        state["door_opened"] = True
    # PROBLEM: continuous per-step bonus
    if state["key_picked_up"] \
       and not state["door_opened"] \
       and "key" in carrying.lower():
        shaped_reward += 0.01  # reward flooding
    # Late-game penalty
    if not state["goal_reached"] \
       and step_count > max_steps * 0.8:
        shaped_reward -= 0.02
    return shaped_reward, state
\end{lstlisting}

\subsection{Iteration 3: Final Refined Function (Final SR: 99\%)}

After three refinement rounds with diagnostic feedback, the function removes continuous bonuses and uses room-aware progress to support stagnation detection:

\begin{lstlisting}[style=python, basicstyle=\ttfamily\tiny]
def reward_fn(obs, action, reward, terminated,
              truncated, info, state):
    shaped_reward = reward
    if "key_picked_up" not in state:
        state["key_picked_up"] = False
    if "door_opened" not in state:
        state["door_opened"] = False
    if "last_agent_pos" not in state:
        state["last_agent_pos"] = (0, 0)
    if "steps_without_position_change" not in state:
        state["steps_without_position_change"] = 0
    if "max_x_before_door" not in state:
        state["max_x_before_door"] = 0
    if "max_x_after_door" not in state:
        state["max_x_after_door"] = 0
    event_text = info.get("event_text", "").lower()
    agent_pos = info.get("agent_pos", (0, 0))
    current_x, current_y = agent_pos
    # ONE-TIME milestone bonuses only
    if not state["key_picked_up"] \
       and "picked up" in event_text \
       and "key" in event_text:
        shaped_reward += 0.2
        state["key_picked_up"] = True
    if not state["door_opened"] \
       and "opened door" in event_text:
        shaped_reward += 0.3
        state["door_opened"] = True
    # Stagnation detection
    if agent_pos != state["last_agent_pos"]:
        state["steps_without_position_change"] = 0
    else:
        state["steps_without_position_change"] += 1
    # Room-aware progress (only resets stagnation state)
    if not state["door_opened"]:
        if current_x > state["max_x_before_door"]:
            state["max_x_before_door"] = current_x
    else:
        if current_x > state["max_x_after_door"]:
            state["max_x_after_door"] = current_x
    # Stagnation penalty (only after 60 stuck steps)
    if state["steps_without_position_change"] > 60 \
       and reward < 0.5:
        shaped_reward -= 0.05
    state["last_agent_pos"] = agent_pos
    return shaped_reward, state
\end{lstlisting}

Key improvements in the final function are: (1)~removing the continuous \texttt{+0.01} holding-key bonus that caused reward flooding, (2)~using room-aware $x$-coordinate progress to reset the stagnation state rather than paying a progress bonus, (3)~replacing the coarse late-game penalty with a stagnation penalty after 60 steps without movement, and (4)~increasing the door-opening bonus from 0.25 to 0.3. These changes resemble the kinds of local refinements a reward engineer might make after inspecting training diagnostics.

\end{document}